\documentclass[runningheads]{llncs}
\usepackage{graphicx}
\usepackage{amsmath,amssymb} 
\usepackage{color}
\usepackage[pagebackref=true,breaklinks=true,letterpaper=true,colorlinks,bookmarks=false]{hyperref}
\usepackage[width=122mm,left=12mm,paperwidth=146mm,height=193mm,top=12mm,paperheight=217mm]{geometry}
\begin{document}

\newcommand\hide[1]{}
\let\bs=\mathbf

\pagestyle{headings}
\mainmatter

\title{StarMap for Category-Agnostic Keypoint and Viewpoint Estimation} 

\titlerunning{StarMap}

\authorrunning{Zhou X., Karpur A., Luo L., Huang Q.}

\author{Xingyi Zhou\inst{1} \and Arjun Karpur\inst{1}
\and Linjie Luo\inst{2} \and Qixing Huang\inst{1}}


\institute{The University of Texas at Austin \and
	Snap Inc. \\
\email{\{zhouxy, akarpur, huangqx\}@cs.utexas.edu, linjie.luo@snap.com}
}

\maketitle

\begin{abstract}
Semantic keypoints provide concise abstractions for a variety of visual understanding tasks. Existing methods define semantic keypoints separately for each category with a fixed number of semantic labels in fixed indices. 
As a result, this keypoint representation is in-feasible when objects have a varying number of parts, 
e.g. chairs with varying number of legs. 
We propose a category-agnostic keypoint representation, which combines a multi-peak heatmap (StarMap) for all the keypoints and their corresponding features as 3D locations in the canonical viewpoint (CanViewFeature) defined for each instance.
Our intuition is that the 3D locations of the keypoints in canonical object views contain rich semantic and compositional information.
Using our flexible representation, we demonstrate competitive performance in keypoint detection and localization compared to category-specific state-of-the-art methods.
Moreover, we show that when augmented with an additional depth channel (DepthMap) to lift the 2D keypoints to 3D,
our representation can achieve state-of-the-art results in viewpoint estimation. 
Finally, we show that our category-agnostic keypoint representation can be generalized to novel categories. 

\end{abstract}

\section{Introduction}

Semantic keypoints, such as joints on a human body or corners on a chair,  provide concise abstractions of visual objects regarding their compositions, shapes, and poses. 
Accurate semantic keypoint detection 
forms the basis for many visual understanding tasks, including human pose estimation~\cite{cao2017realtime,newell2016stacked,pavlakos2017coarse,zhou2016deep}, 
hand pose estimation~\cite{yuan20173d,zhou2016model}, 
viewpoint estimation~\cite{pavlakos20176,tulsiani2015viewpoints},
feature matching~\cite{long2014convnets}, fine-grained image classification~\cite{zhang2014part}, and 3D reconstruction~\cite{drcTulsiani17,marrnet,shapesKarTCM15,cmrKanazawa18}.

Existing methods define a fixed number of semantic keypoints for each object category in isolation~\cite{tulsiani2015viewpoints,pavlakos20176,wu2016single,newell2016stacked}. A standard approach is to allocate a heatmap channel for each keypoint. Or in other words, keypoints are inferred as separate heat maps according to their \emph{encoding order}. 
This approach, however, is not suitable when objects have a varying number of parts, e.g. chairs with varying numbers of legs. 
The approach is even more limiting when we want to share and use keypoint labels of multiple different categories. 
In fact, keypoints of different categories do share rich compositional similarities. 
For instance, chairs and tables may share the same configuration of legs, and motorcycles and bicycles all contain wheels. 
Category-specific keypoint encodings fail to capture both the intra-category part variations and the inter-category part similarities.

\begin{figure}[t]
\centering
\includegraphics[width=0.8\linewidth]{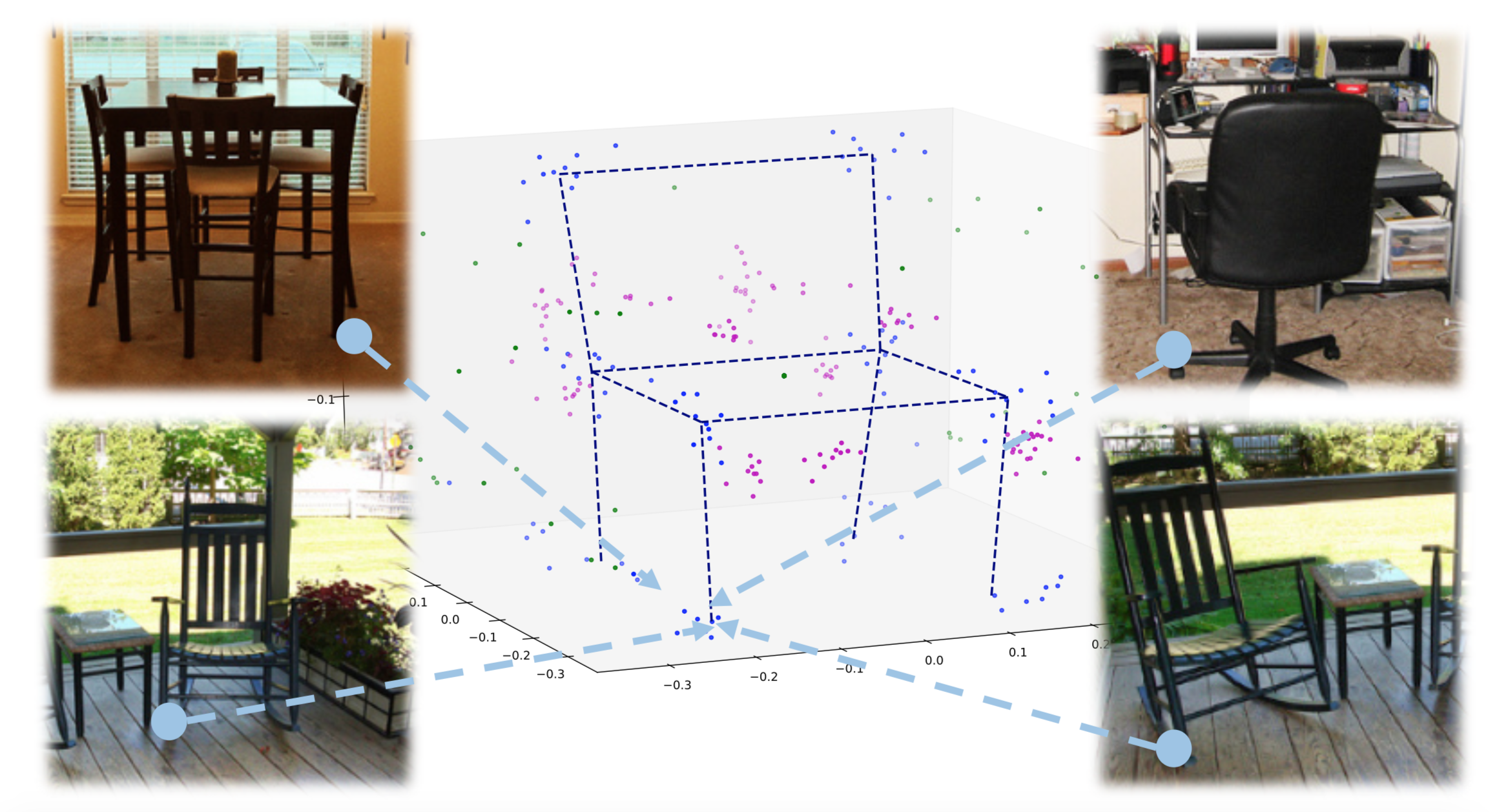}
\caption{Illustration of Canonical View Semantic Feature. It is shared across all object categories. We show 2 categories: chair (in blue) and table (in green). For the left frontal leg of chair on bottom left, it has i) the same CanViewFeature with the same chair keypoint from a different viewpoint (bottom right), ii) similar feature with another chair instance's corresponding keypoint (top right), and iii) similar feature with left frontal leg from a table(top left). We \emph{Can View} this feature in 3D space (middle).}
\label{fig:canview}
\end{figure}

In this paper, we propose a novel, category-agnostic keypoint representation.
Our representation consists of two components: 1) a single channel, multi-peak heatmap, termed \emph{StarMap}, for all keypoints of all objects; 
and 2) their respective feature (Fig.~\ref{fig:canview}), termed \emph{CanViewFeature}, which is defined as the 3D locations in a normalized canonical object view (or a world coordinte system). 
Specifically, StarMap combines the separate keypoint heat maps in previous approaches~\cite{tulsiani2015viewpoints,pavlakos20176} into a single heat map, and thus unifies the detection of different keypoints. 
CanViewFeature provides semantic discrimination between keypoints, i.e., through their locations in the normalized canonical object view.
One intuition behind this representation is that the distribution of keypoints' 3D locations in the canonical object view encodes rich semantic and compositional information. For example, the locations of all legs are close to the ground, and they are below the seats.
Our representation can be obtained via \emph{supervised} training on any standard datasets with 3D viewpoint annotations, such as Pascal3D+~\cite{xiang2014beyond} and ObjectNet3D~\cite{xiang2016objectnet3d}.

Our representation provides the flexibility to represent varying numbers of keypoints across different categories by eliminting the hard-encoding of keypoints. Additionally, we demonstrate that our representation can still achieve competitive results in keypoint detection and localization compared to the state-of-the-art category-specific approaches~\cite{long2014convnets,tulsiani2015viewpoints} (Sec~\ref{Subsection:Keypoint:Detection:Classification}) by using simple nearest neighbor association on the category-level keypoint templates.

One direct application of our representation is viewpoint estimation~\cite{tulsiani2015viewpoints,su2015render,mousavian20173d}, which can be achieved by solving a perspective-n-points (PnP)~\cite{lepetit2009epnp} problem to align the \emph{CanViewFeature} with the \emph{StarMap}. 
Further, we observed considerable performance gains in this task by augmenting the \emph{StarMap} with an additional depth channel (\emph{DepthMap}) to lift the 2D image coordinates into 3D. 
We report state-of-the-art performance compared to previous viewpoint estimation methods~\cite{su2015render,pavlakos20176,mousavian20173d,tulsiani2015viewpoints} with ablation studies on each component.
Finally, we show our method works well when applied to unseen categories. 
Full code is publicly available at 
\url{https://github.com/xingyizhou/StarMap}. 

 \section{Related Works}

\textbf{Keypoint estimation.}
Keypoint estimation, especially human joint estimation~\cite{toshev2014deeppose,tompson2015efficient,newell2016stacked,cao2017realtime,Zhou_2017_ICCV} and rigid object keypoint estimation~\cite{wu2016single,zhou2017unsupervised}, is a widely studied problem in computer vision.
In the simplest case, a 2D/3D keypoint can be represented by a 2/3-dimension vector and learned by supervised regression.
Toshev et al.~\cite{toshev2014deeppose} first trained a deep neural network for 2D human pose regression
and Li et al.~\cite{li20143d} extended this approach to 3D.
Starting from Tompson et al.~\cite{tompson2014joint}, the heatmap representation has dominated the 2D keypoint estimation community and has achieved great success in both 2D human pose estimation~\cite{newell2016stacked,wei2016convolutional,yang2017learning} and \emph{single category} man-made object keypoint detection~\cite{wu2016single,marrnet}.
Recently, the heatmap representation has been generalized in various different directions.
Cao et al.~\cite{cao2017realtime} and Newell et al.~\cite{newell2017associative} extended the single peak heatmap (for single keypoint detection) to a multi-peak heatmap where each peak is one instance of a \emph{specific type} of keypoint, enabling bottom-up, multi-person pose estimation.
Pavlakos et al.~\cite{pavlakos2017coarse} lifted the 2D pixel heatmap to a 3D voxel heatmap, resulting in an end-to-end 3D human pose estimation system.
Tulsiani et al.~\cite{tulsiani2015viewpoints} and Pavlakos et al.~\cite{pavlakos20176} stacked keypoint heatmaps from different object categories together for multi-category object keypoint estimation.
Despite good performance gained by these approaches, they share a common limitation:
each heatmap is only trained for a \emph{specific keypoint type} from a specific object.
Learning each keypoint individually not only ignores the intra-category variations or inter-category similarities, but also makes the representation inherently impossible to be generalized to unknown keypoint configurations for novel categories.

\textbf{Viewpoint estimation.}
Viewpoint estimation, i.e., estimating an object's orientation in a given frame, is a practical problem in computer vision and robotics~\cite{kendall2015posenet,pavlakos20176}.
It has been well explored by traditional techniques that solve for transformations between corresponding points in the world and image views; this is known as the Perspective-n-Point  Problem~\cite{lepetit2009epnp,lu2000fast}.
Lately, viewpoint estimation accuracy and utility have been greatly improved in the deep learning era.
Tulsiani et al.~\cite{tulsiani2015viewpoints} introduced viewpoint estimation as a bin classification problem for each viewing angle (azimuth, elevation and in-plane rotation). 
Mousavian et al.~\cite{mousavian20173d} augmented the bin classification scheme by adding regression offsets within each bin so that predictions could be more fine-grained. 
Szeto et al.~\cite{szeto2017click} used annotated keypoints as additional input to further improve bin classification.
To combat scarcity of training data and generic features, Su et al.~\cite{su2015render} proposed to synthesize images with known 3D viewpoint annotations and proposed a geometry-aware loss to further boost the estimation performance.
Recently, Pavlakos et al.~\cite{pavlakos20176} proposed to use detected \emph{semantic} keypoint followed by a PnP algorithm~\cite{lepetit2009epnp} to solve for the resulting viewpoint matrix and achieved state-of-the-art results. 
However, this method relies on category-specific keypoint annotation and is not generalizable. 
On the contrary, our approach is both accurate and category-agnostic, by utilizing category-agnostic keypoints. 

\textbf{General keypoint detection.}
There are several related concepts similar to our general semantic keypoint. 
The most well-known one is the SIFT descriptor~\cite{lowe2004distinctive}, which aims to detect a large 
number of \emph{interest} points based on local and \emph{low level} image statistics. 
Also, the heatmap representation has been used in saliency detection~\cite{huang2015salicon} and visual attention~\cite{xu2015show}, which detects a \emph{region} of image
which is ``important'' in the context.
Similarly, Altwaijry et al.~\cite{altwaijry2016learning} used the heatmap representation to detect a set of points that is useful for feature matching. 
The key difference between our keypoint and the above concepts is that their keypoints do not contain semantic meanings and are not annotated by humans, making them less useful in high level vision tasks such as pose estimation.  

To our best knowledge, we are the first to propose a category-agnostic keypoint representation and show that it is directly applicable to viewpoint estimation.

\section{Approach}

\begin{figure}[t]
\centering
\includegraphics[width=0.9\linewidth]{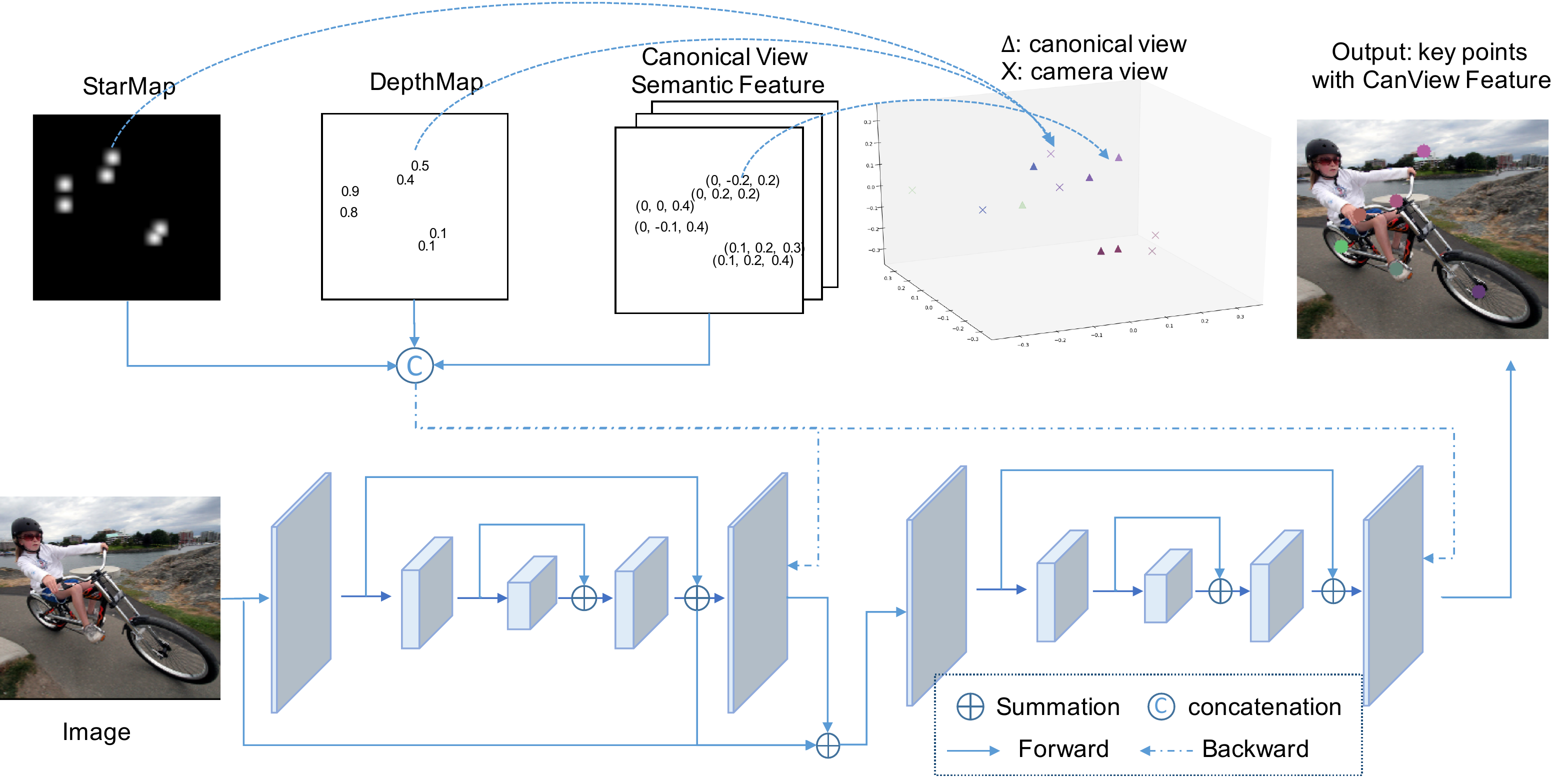}
\caption{Illustration of our framework. For an input image, our network predicts three component: StarMap, Canonical View Feature, and DepthMap. Varying number of keypoints are extracted at the peak location of StarMap and their Depth and CanViewFeature can be accessed at the corresponding channels.}
\label{fig:framework}
\end{figure}

In this section, we describe our approach for learning a category-agnostic keypoint representation from a single RGB image. 
We begin with describing the representation in Section~\ref{Section:Problem:Statement}.
We then introduce how to learn this representation in Section~\ref{Section:Training:Data:Preparation}.
Finally, we show a direct application of our representation in viewpoint estimation in Section~\ref{Section:Pose:Estimation}.

\subsection{Category-agnostic keypoint representation}
\label{Section:Problem:Statement}

A desired general purpose keypoint representation should 
be both adaptive (i.e., should be able to represent different content of different visual objects) and semantically meaningful (i.e., should convey certain semantic information for downstream applications). 

So far the most widely used keypoint representation is the category specific stacked keypoint vector~\cite{toshev2014deeppose}, which represents object keypoints by a $N \times D$ vector ($N$ for number of keypoints and $D$ for dimensions), or multi-channel heatmaps~\cite{tompson2014joint,newell2016stacked}, which associate each channel with one specific keypoint on a specific object category, e.g., $16$-channel heatmaps for human~\cite{tompson2014joint,newell2016stacked}, $10$-channel heatmaps for chair~\cite{wu2016single}.
Although these representations are certainly semantically meaningful (e.g., the first channel of human heatmaps is the left ankle), it does not satisfy the adaptive property, e.g., chairs with legged bases and swivel bases cannot be learned together due to varying number of keypoints.
As a result, they can not be considered as the same category based on their different keypoint configurations.
To generalize heatmaps to multiple categories, 
a popular approach is to stack all heatmaps from all categories~\cite{tulsiani2015viewpoints,pavlakos20176} (resulting in $\sum{N_c}$ output channels, where $N_c$ is the number of keypoints of category $c$).
In such a representation, keypoints from different objects are completely separated,
e.g. seat corners from swivel chairs are irrelevant to seat corners from chairs. To merge keypoints from different objects, one has to establish consistent correspondences~\cite{zhou2016learning} between different keypoints across multiple categories, which is difficult or sometimes impossible.

In this paper, we introduce a hybrid representation that meets all desired properties. 
As illustrated in Figure~\ref{fig:framework}, our hybrid representation consists of three components, \emph{StarMap}, \emph{CanViewFeature} and \emph{DepthMap}. 
In particular, StarMap specifies the image coordinates of keypoints where the number of keypoints can \emph{vary} across different categories; 
CanViewFeature specifies the 3D locations of keypoints in a canonical coordinate system, which provide an identity for each keypoint; 
DepthMap lifts 2D keypoints into 3D. As we will see later, it enhances the performance of using this representation for the application of viewpoint estimation. 
Now we describe each component in more details.

\noindent\textbf{StarMap.} As shown in Figure~\ref{fig:framework} (top left), StarMap is a single channel heatmap whose local maximums encode the image locations of the underlying points. 
It is motivated by the success of using one heatmap to encode occurrences of one keypoint on multiple persons~\cite{cao2017realtime,newell2017associative}. 
In our setting, we generalize the idea to encode \emph{all} keypoints of each object.
This is in contrast to~\cite{cao2017realtime,newell2017associative}, which use multi-peak heatmaps to detect multiple \emph{instances} of the \emph{same} specific keypoint. In our implementation, given a heatmap, we extract the corresponding keypoints by detecting all local maximums, with respect to the 8-ring neighborhood whose values are above $0.05$. 

When comparing multi-channel heatmaps and a single channel heatmap, one intuition is that multi-channel heatmaps, which are category-specific and keypoint-specific representations, lead to better accuracy.
However, as we will see later, using a single channel allows us to train the representation from bigger training data (multiple categories), leading to an overall better keypoint predictor. 
We also argue that a single-channel representation (1 channel vs 100+ channels on Pascal3D+~\cite{xiang2014beyond}) is favored when computational and memory resources are limited.
On the other hand, \emph{StarMap} alone does not provide the semantic meaning of each detected point. 
This drawback motivates the second component of our hybrid keypoint representation.

\noindent\textbf{CanViewFeature.} CanViewFeature collects the 3D locations of the keypoints in the canonical view. 
In our implementation, we allocate three channels for \emph{CanViewFeature}. Specifically, after detecting a keypoint (peak) in \emph{StarMap}, the values of these three channels at the corresponding pixel specify the 3D location in the canonical coordinate system.
The design of \emph{CanViewFeature} is motivated from recent works on embedding visual objects into latent spaces~\cite{taylor2012vitruvian,DBLP:conf/cvpr/WeiHCVL16}.
Such latent spaces provide a shared platform for comparing and linking different visual objects.
Our representation shares the same abstract idea, yet we make the embedding \emph{explicit} in 3D  (where we \emph{can view} the learned representation) and
learnable in a \emph{supervised} manner.
This enables additional applications such as viewpoint estimation, as we will discuss later. 
When considering the space of keypoint configurations in the canonical space, 
it is easy to find that the feature is invariant to object pose and image appearance (scale, translation, rotation, lighting), little-variance 
to object shape (e.g., left frontal wheels from different cars are always in the left frontal area), and little variance to object category (e.g., frontal wheels from different categories are always in bottom frontal area).

Although \emph{CanViewFeature} only provides 3D locations, we can leverage this to classify the keypoints,
by using nearest neighbor association on the category-level keypoint templates. 

\noindent\textbf{DepthMap.} \emph{CanViewFeature} and \emph{StarMap} are related to each other via a similarity transform (rotation, translation, scaling) and a perspective projection. 
It is certainly possible to solve a non-linear optimization problem to recover the underlying similarity transform. 
However, since the network predictions are not perfect, we found that this approach leads to sub-optimal results. 

To stabilize this process and make the relation even simpler, we augment \emph{StarMap} with one additional channel called \emph{DepthMap}. 
The encoding is the same as \emph{CanViewFeature}. 
More precisely, we first extract keypoints at peak locations and then access the corresponding pixels to obtain the depth values.
When the camera intrinsic parameters are present, we use them to convert image coordinates and depth value into the true 3D location of the corresponding pixel. 
Otherwise, we assume weak-perspective projection, and directly use the image coordinates and depth value as an approximation of the underlying 3D location.

\label{Section:Representation:Encoding}

\subsection{Learning Hybrid Keypoint Representation}
\label{Section:Training:Data:Preparation}

\noindent\textbf{Data preparation.} 
Training our hybrid representation requires annotations of 2D keypoints, their corresponding depths, and their corresponding 3D locations in the canonical view.
We remark that such training data is feasible to obtain and publicly available~\cite{xiang2014beyond,xiang2016objectnet3d}. 
2D keypoint annotations per image are straightforward to retrieve~\cite{papadopoulos2017extreme} and thus widely available~\cite{lin2014microsoft,andriluka14cvpr,bourdev2010detecting}.
Also, annotating 3D keypoints of a CAD model~\cite{yi2016scalable} is not a hard task, given an interactive 3D UI such as MeshLab~\cite{LocalChapterEvents:ItalChap:ItalianChapConf2008:129-136}. 
The canonical view of a CAD model is defined as the front view of an object with the largest 3D bounding box dimension scaled to $[-0.5, 0.5]$ (meaning it is zero centered). 
Note that just a few 3D CAD models need to be annotated for each category (about 10 per category), because keypoint configuration variation is orders of magnitude smaller than the image appearance variation. 
Given a collection of images and a small set of CAD models of the corresponding categories, a human annotator is asked to select the closest CAD model to the image's content, as done in Pascal3D+ and ObjectNet3D~\cite{xiang2014beyond,xiang2016objectnet3d}.
A coarse viewpoint is also annotated by manually dragging the selected CAD model to align the image appearance. 
In summary, all the annotations required to train our hybrid representation are relatively easy to acquire. 
We refer to~\cite{xiang2014beyond,xiang2016objectnet3d} for more details on how to annotate such data. 

We now describe how we calculate the depth annotation.
Ideally, the transformation between the canonical view and image pixel coordinate is a full-perspective camera model:
\begin{equation}
s [u \  v \  1]^T = \mathcal{A} [R|t] [\overline{x} \  \overline{y} \  \overline{z} \  1]^T, s.t., R^T R = I
\label{eq:perspective}
\end{equation}
where $\mathcal{A}$ describes intrinsic camera parameters, $(u, v)$ is the 2D keypoint location in the image coordinate system, $(\overline{x}, \overline{y}, \overline{z})$ is the 3D location in canonical coordinate system. $R$, $t$, and $s$ are the rotation matrix (i.e. viewpoint), translation vector, and scale factor, respectively. 
However, the camera intrinsic parameters are most likely unavailable in testing scenarios.
In those cases, a weak-perspective camera model is often applied to \emph{approximate} the 3D-to-2D transformation for keypoint estimation~\cite{Zhou_2017_ICCV,pavlakos20176}, by changing Eq.~\ref{eq:perspective} to
\begin{equation}
s [u - c_x\  v - c_y \  d]^T = [R|t] [\overline{x} \  \overline{y} \  \overline{z} \  1]^T, s.t., R^T R = I
\label{eq:perspective_2}
\end{equation}
where $(u, v)$ specifies the location of the keypoint, $d$ is its associated depth, and $(c_x, c_y)$ denotes the center of the image.

Letting $[{x}, {y}, {z}] = [R|t] [\overline{x}, \overline{y}, \overline{z}, 1]^T$ be the transformed 3D keypoints in the metric space, we have $[u, v, d] = [{x} / s + c_x, {y} / s + c_y, {z} / s]$ (with unknown $s$), which transforms one point from the 3D metric space to the 2D pixel space with an augmented depth value $d$.
In training, let $N_c$ be the number of keypoints in category $c$. Both the viewpoint transformation matrix $[R|t]$ and the canonical points $\{\overline{x}_i, \overline{y}_i, \overline{z}_i\}_{i = 1} ^ {N_c}$ are known, and we can calculate the rotated keypoints $\{{x}_i, {y}_i, {z}_i\}_{i = 1}^{N_c}$. 
Moreover, the corresponding 2D keypoints $\{(u_i, v_i)\}_{i = 1}^{N_c}$ are known, so we can simply solve the scale factor $s$ by aligning the $(u, v)$ and $({x}, {y})$ plane bounding box size:
$s = \frac{max(max_i({x}_i) - min_i({x}_i), max_i({y}_i) - min_i({y}_i))}{max(max_i(u_i) - min_i(u_i), max_i(v_i) - min_i(v_i))}$, which gives rise to the underlying depth value.

\noindent\textbf{Network training.} 
As described above, we have full supervision for all of our 3 output components. 
Training is done as a supervised heatmap regression, i.e., we minimize the $L2$ distance between the output 5-channel heatmap and their ground truth. 
Note that for \emph{CanViewFeature} and \emph{DepthMap}, we only care about the output at peak locations.
Following~\cite{newell2017pixels,newell2017associative}, 
we ignore the non-peak output locations rather than forcing them to be zero.
This can be simply implemented by multiplying a mask matrix to both the network output and ground truth and then using a standard $L2$ loss. 

\noindent\textbf{Implementation details.}
Our implementation is done in the PyTorch framework. 
We use a 2-stacks HourglassNetwork~\cite{newell2016stacked}, which is the state-of-the-art architecture for 2D human pose estimation~\cite{andriluka14cvpr}. 
We trained our network using curriculum learning, i.e., we first train the network with only \emph{StarMap} output for 90 epochs and then fine-tune the network with the \emph{CanViewFeature} followed by \emph{DepthMap} supervision for additional 90 epochs each. 
The whole training stages took about 2 days on one GTX 1080 TI GPU. 
All the hyper-parameters are set to the default values in the original Hourglass implementation~\cite{newell2016stacked}.

\subsection{Application in Viewpoint Estimation}
\label{Section:Pose:Estimation}

The output of our approach (StarMap, DepthMap and CanViewFeature) can directly be used to estimate the viewpoint of the input image with respect to the canonical view (i.e., camera pose estimation). Specifically, Let ${\mathbf{p}}_i = (u_i-c_x, v_i-c_y,d_i)$ be the un-normalized 3D coordinate of keypoint $p_i$, where ($c_x$, $c_y$) is the image center. Let $\mathbf{q}_i$ be its counterpart in the canonical view. With $w_i\in [0,1]$ we denote this keypoint's value on the heatmap, which indicates a confidence score. We solve for a similarity transformation between the image coordinate system and world coordinate system that is parameterized by a scalar $s\in \mathbb{R}^{+}$, a rotation $R \in SO(3)$, and a translation $\bs{t}$. This is done by minimizing the following objective function:
\begin{equation}
s^{\star}, R^{\star}, \mathbf{t}^{\star} = \underset{s, R, \mathbf{t}}{\textup{argmin}}\ \sum\limits_{i=1}^{N_I} w_i\|sR{\mathbf{p}}_i + \mathbf{t}-\mathbf{q}_i\|^2.
\label{Eq:Pose:Estimation}
\end{equation}
Note that (\ref{Eq:Pose:Estimation}) admits an explicit solution as described in~\cite{horn1987closed}, which we include here for completeness. The optimal rotation is given by
\begin{equation}
R^{\star} = U\textup{diag}(1,1,\textup{sign}(M))V^{T},\qquad M:= \sum\limits_{i=1}^{N_I}w_i (\mathbf{p}_i-\overline{\mathbf{p}})(\mathbf{q}_i-\overline{\mathbf{q}})
\label{Eq:Pose:Solution}
\end{equation}
where $U\Sigma V^{T} = M$ is the SVD and $\overline{\mathbf{p}}$, $\overline{\mathbf{q}}$ are the mean of $\mathbf{p}_i$, $\mathbf{q}_i$. 

\section{Experiments}

In this section, we perform experimental evaluations on the proposed hybrid keypoint representation. We begin with describing the experimental setup in Section~\ref{Subsection:Experimental:Setup}. We then evaluate the accuracy of our keypoint detector and the application in viewpoint estimation in Section~\ref{Subsection:Keypoint:Detection:Classification} and Section~\ref{Subsection:Viewpoint:Estimation}, respectively. We then present advanced analysis of our hybrid keypoint representation in Section~\ref{Subsection:Advanced:Analysis}. 
Finally, we show that our category-agnostic keypoint representation can be extended to novel categories in Section~\ref{Subsection:PoseInduction}.
Table~\ref{table:demo} collect some qualitative results, and more results are deferred to the supplementary material.

\subsection{Experimental Setup}
\label{Subsection:Experimental:Setup}

We use Pascal3D+~\cite{xiang2014beyond} as our major evaluation benchmark.  This dataset contains 12 man-made object categories with 2K to 4K images per category. We make use of the following annotations in our training:
object bounding box, category-specific 2D keypoints (annotations from~\cite{bourdev2010detecting}), approximate 3D CAD model of the object, viewpoint of the image, and category-specific 3D keypoint annotations (corresponds with the 2D keypoint configuration) in the canonical coordinate system defined on each CAD model.
Following~\cite{tulsiani2015viewpoints,su2015render}, evaluation is done on the subset of the validation set that is non-truncated and non-occluded, which contains $2113$ samples in total. 
As the evaluation protocols and baseline approaches vary across different tasks, we will describe them for each specific set of evaluations.

\subsection{Keypoint Localization and Classification}
\label{Subsection:Keypoint:Detection:Classification}

We first evaluate our method on the keypoint estimation task, which specifies the locations of the predicted keypoints. 
Since keypoint locations alone do not carry the identities of each keypoint and cannot be used as identity-specific evaluation, we perform the evaluation by using two protocols --
namely, with identification inferred from our learned CanViewFeature or with oracle assigned identification.
Specifically, for the first protocol, for each category, we calculate the mean of the locations of each keypoint in the world coordinate system among all CAD models and use this as the \emph{category-level} template. 
We then associate each keypoint with the ID of its nearest mean annotated keypoint in the template.
For the second protocol, we assume a perfect ID assignment (or keypoint classification) by assigning the output keypoint ID as the closest annotation (in image coordinates).
The second protocol can also be thought of as randomly perturbing the annotated keypoint order and picking the best one. 
Following the conventions~\cite{long2014convnets,tulsiani2015viewpoints}, we use PCK($\alpha = 0.1$), or Percentage of Correct Keypoints, as the evaluation metric. PCK considers a keypoint to be correct if its $L2$ 2D pixel distance from the ground truth keypoint location is less than $0.1 \times max(h, w)$, where $h$ and $w$ are the object's bounding box dimensions.

\begin{center}
\begin{table*}[t]
\scriptsize
\centering
\begin{tabular}{l|cccccccccccc|c}
\hline
PCK($\alpha=0.1$) & {{}aero} & {{}bike} & {{}boat} & {{}bottle} & {{}bus} & {{}car} & {{}chair} & {{}table} & {{}mbike} & {{}sofa} & {{}train} & { {}tv} & {mean} \\
\hline
{{}{Long. \cite{long2014convnets}}} & 53.7 & 60.9 & 33.8 & 72.9 & 70.4 & 55.7 & 18.5 & 22.9 & 52.9 & 38.3 & 53.3 & 49.2 & 48.5 \\
{Tulsiani.~\cite{tulsiani2015viewpoints}} & 66.0 & 77.8 & 52.1 & 83.8 & 88.7 & 81.3 & 65.0.& 47.3 & 68.3 & 58.8 & 72.0 & 65.1 & 68.8 \\
{Pavlakos.~\cite{pavlakos20176}} & 84.1 & 86.9 & 62.3 & 87.4 & 96.0 & 93.4 & 76.0 & N/A & N/A & 78.0 & 58.4 & 84.8 & 82.5 \\
Ours & 75.2 & 83.2 & 54.8 & 87.0 & 94.4 & 90.0 & 75.4 & 58.0 & 68.8 & 79.8 & 54.0 & 85.8 & 78.6 \\
\hline
\hline
{Pavlakos.~\cite{pavlakos20176} Oracle Id} & 92.3 & 93.0 & 79.6 & 89.3 & 97.8 & 96.7 & 83.9 & N/A & N/A & 85.1 & 73.3 & 88.5 & 89.0 \\
Ours Oracle Id & 93.1 & 92.6 & 84.1 & 92.4 & 98.4 & 96.0 & 91.7 & 90.0 & 90.1 & 89.7 & 83.0 & 95.2 & 92.2 \\
\hline
\end{tabular}
\caption{2D Keypoint Localization Results. The results are shown in PCK($\alpha = 0.1$). Top: our result with nearest canonical feature as keypoint identification. Bottom: results with oracle keypoint identification.}
\label{table:kpsGtEval}
\end{table*}
\end{center}

The keypoint localization and classification results are shown in Table~\ref{table:kpsGtEval}.
We show 3 state-of-the-art methods~\cite{long2014convnets,tulsiani2015viewpoints,pavlakos20176} for \emph{category-specific} keypoint localization for comparison. 
The evaluation of~\cite{pavlakos20176} is done by ourselves based on their published model. 
For the first protocol, 
our result of $78.6\%$ mean PCK($\alpha = 0.1$) is marginally better than the state-of-the-arts in 2014~\cite{long2014convnets,tulsiani2015viewpoints}, probably because we used a more up-to-date HourglassNetwork~\cite{newell2016stacked}. 
Our performance is slightly worse than~\cite{pavlakos20176}, who uses the same Hourglass architecture but with stacked category-specific channels output ($\sum_{c} N_c$ output channels in total), which is expected.
This is due to the error caused by incorrect keypoint ID association. 
We emphasize that all counterpart methods are category-specific, thus requiring ground truth object category as input while ours is general. 

The second protocol (Bottom of Table~\ref{table:kpsGtEval}) factors out the error caused by incorrect keypoint ID association.
For a fair comparison, we also allow~\cite{pavlakos20176} to change its output order with the oracle nearest location (to eliminate the common left-right flip error~\cite{Ronchi_2017_ICCV}).  
We can see our score is $92.2\%$, which is $3.2\%$ higher than that of Pavlakos et al~\cite{pavlakos20176}. 
This is quite encouraging since our approach is designed to be a general purpose keypoint predictor. 
This result shows that it is advantageous to train a unified network to predict keypoint locations, as this allows to train a single network with more relevant training data.

\subsection{Viewpoint Estimation}
\label{Subsection:Viewpoint:Estimation}

Some qualitative results are shown in Table.~\ref{table:demo}, and more results can be found in the supplementary material.

As a direct application, we evaluate our hybrid representation on the task of viewpoint estimation. The objective of viewpoint estimation is to predict the azimuth ($a$), elevation ($e$), and in-plane rotation ($\theta$) of the image object with respect to the world coordinate system. In our experiment, we follow the conventions~\cite{tulsiani2015viewpoints,su2015render} by measuring the angle between the predicted rotation vector and the ground truth rotation vector:
$
\Delta(R_{pred}, R_{gt}) = \frac{||logm(R_{pred}^TR_{gt})||_{\mathcal{F}}}{\sqrt{2}},
$
where $R = R_{Z}(\theta) R_X(e - \pi / 2) R_Z(- a)$ transforms the viewpoint representation $(a, e, \theta)$ into a rotation matrix. Here $R_X$, $R_Y$ and $R_Z$ are rotations along $X$, $Y$ and $Z$ axis, respectively. 

We consider two metrics that are commonly applied in the literature~\cite{tulsiani2015viewpoints,pavlakos20176,mousavian20173d,su2015render}, namely, Median Error, which is the median of the rotation angle error, and Accuracy at $\theta$, which is the percentage of keypoints whose error is less than $\theta$. We use $\theta = \frac{\pi}{6}$, which is a default setting in the literature.

\begin{table*}[t]
\scriptsize
\centering
\resizebox{\columnwidth}{!}{%
\begin{tabular}{c|cccccccccccc|c}
\hline
& aero &bike& boat& bottle& bus& car& chair& table& mbike& sofa& train& tv & mean\\
\hline
$\mathit{MedErr}$(Tulsiani~\cite{tulsiani2015viewpoints}) & 13.8 &17.7& { 21.3}& 12.9& 5.8& 9.1& 14.8& 15.2& 14.7& 13.7& 8.7& 15.4  & 13.6\\
$\mathit{MedErr}$(Pavlakos~\cite{pavlakos20176}) & { 8.0}& 13.4& 40.7& 11.7& { 2.0}& { 5.5} & { 10.4} & N/A & N/A & { 9.6} & 8.3 & 32.9 & N/A\\
$\mathit{MedErr}$(Mousavian~\cite{mousavian20173d}) &  13.6 & { 12.5} & 22.8 & { 8.3} & 3.1&  5.8 & 11.9 & { 12.5} & { 12.3} & 12.8 &{ 6.3}& { 11.9} & { 11.1}\\
$\mathit{MedErr}$(Su~\cite{su2015render}) &  15.4 & 14.8 & 25.6 & 9.3 & 3.6 &  6.0 & 9.7 & 10.8 & 16.7 & 9.5 & 6.1 & 12.6 & 11.7\\
$\mathit{MedErr}$(Mahendran~\cite{mahendran2017joint}) & 14.2 & 18.7 & 27.2 & 9.5 & 3.0 &  6.9 & 15.8 & 14.4 & 16.4 & 10.7 & 6.6 & 14.3 & 13.1\\
$\mathit{MedErr}$(Res18-General) &  14.3  & 16.7  & 26.9  & 13.2 & 5.8 & 8.8  & 17.7 &  26.7 & 15.7  & 14.4  & 8.8 & 16.2  & 13.3\\
$\mathit{MedErr}$(Res18-Specific) &  14.7  & 15.8  & 25.6  & 13.1 & 5.7 & 8.6  & 16.3 &  18.1 & 15.1  & 13.8  & 8.2 & 14.1  & 12.8\\
$\mathit{MedErr}$(PnP) &  9.5  & 14.0  & 43.6  & 9.9 & 3.3 & 6.6  & 11.4  &  64.9 & 14.3  & 11.5  & 7.7 & 21.8  & 11.2 \\
$\mathit{MedErr}$(Ours) &  10.1  & 14.5  & 30.0  & 9.1 & 3.1 & 6.5  & 11.0  &  23.7 & 14.1  & 11.1  & 7.4 & 13.0  & {\bf 10.4}\\
\hline
$\mathit{Acc}_\frac{\pi}{6}$(Tulsiani~\cite{tulsiani2015viewpoints})& { 0.81}& 0.77& { 0.59}& { 0.93}& { 0.98}& 0.89& { 0.80}& 0.62& { 0.88}& { 0.82}& 0.80& 0.80& 0.8075\\
$\mathit{Acc}_\frac{\pi}{6}$(Pavlakos~\cite{pavlakos20176})& 0.81& 0.78& 0.44& 0.79 &0.96& 0.90& 0.80& N/A & N/A & 0.74& 0.79& 0.66 & N/A \\
$\mathit{Acc}_\frac{\pi}{6}$(Mousavian~\cite{mousavian20173d})& 0.78& {0.83} & 0.57& { 0.93} &0.94& { 0.90}& { 0.80}& { 0.68} & 0.86& { 0.82}& { 0.82}& { 0.85} & { 0.8103} \\
$\mathit{Acc}_\frac{\pi}{6}$(Su~\cite{su2015render})& 0.74 & 0.83 & 0.52 & 0.91 & 0.91 & 0.88 & 0.86 & 0.73 & 0.78 & 0.90 & 0.86 & 0.92 & 0.82  \\
$\mathit{Acc}_\frac{\pi}{6}$(Res18-General)& 0.79 & 0.75 & 0.53 & 0.90 & 0.96 & 0.93 & 0.62 & 0.57 & 0.85 & 0.82 & 0.81 & 0.77 & 0.7875 \\
$\mathit{Acc}_\frac{\pi}{6}$(Res18-Specific)& 0.79& 0.77 & 0.54 & 0.93 & 0.95 & 0.93 & 0.75 & 0.57 & 0.84 & 0.79 & 0.81 & 0.84 & 0.8121 \\
$\mathit{Acc}_\frac{\pi}{6}$(PnP)& 0.80 & 0.70 & 0.37 & 0.88 & 0.94 & 0.86 & 0.76 & 0.48 & 0.80 & 0.92 & 0.74 & 0.57 & 0.7416  \\
$\mathit{Acc}_\frac{\pi}{6}$(Ours)& 0.82& 0.86 & 0.50 & 0.92 & 0.97 & 0.92 & 0.79 & 0.62 & 0.88 & 0.92 & 0.77 & 0.83 & {\bf 0.8225}  \\
\hline
$\mathit{Acc}_\frac{\pi}{18}$(Res18-General)& 0.28 & 0.18 & 0.17 & 0.27 & 0.82 & 0.61 & 0.23 & 0.33 & 0.18 & 0.15 & 0.61 & 0.27 & 0.3502 \\
$\mathit{Acc}_\frac{\pi}{18}$(Res18-Specific)& 0.29& 0.21 & 0.21 & 0.30 & 0.86 & 0.62 & 0.28 & 0.33 & 0.21 & 0.18 & 0.59 & 0.30 & 0.3777 \\
$\mathit{Acc}_\frac{\pi}{18}$(PnP)& 0.52 & 0.36 & 0.13 & 0.50 & 0.83 & 0.65 & 0.48 & 0.29 & 0.31 & 0.44 & 0.61 & 0.27 & 0.4643\\
$\mathit{Acc}_\frac{\pi}{18}$(Ours)& 0.49& 0.34 & 0.14 & 0.56 & 0.89 & 0.68 & 0.45 & 0.29 & 0.28 & 0.46 & 0.58 & 0.37 & {\bf 0.4818}\\
\hline
\end{tabular}
}
\caption{Viewpoint Estimation on Pascal3D+~\cite{xiang2014beyond}. We compare our results with the state-of-the-arts and baselines. The results are shown in Median Error (lower better) and Accuracy (higher better).}
\label{tab:pascal_viewpoint}
\end{table*}

A popular approach for solving viewpoint estimation is to cast the problem as bin classification by discretiziing the space of $(a, e, \theta)$~\cite{tulsiani2015viewpoints,mousavian20173d,su2015render,mahendran2017joint}. Since network architecture governs the performance of a neural network, we re-train the baseline models~\cite{tulsiani2015viewpoints} with more modern network architectures~\cite{he2016deep}.
We implemented a ResNet18 (\textbf{Res18-Specific}) with the same hyper-parameters as~\cite{tulsiani2015viewpoints} (we also tried VGG~\cite{simonyan2014very} or ResNet50~\cite{he2016deep} but observed very similar or worse performance).

We also want to remark that although viewpoint estimation itself is not a category-specific task, all the studied preview works have used a category-specific formulation, e.g., use \emph{separate} last-layer bin classifiers for each category, resulting in $3 \times N_{categories} \times N_{bins}$ output units~\cite{tulsiani2015pose}. 
We also provide a general $3 \times N_{bins}$ viewpoint estimator as a baseline (\textbf{Res18-General}).

Table~\ref{tab:pascal_viewpoint} compares our approach with previous techniques. Our method outperforms all previous methods and baselines in both testing metrics. Specifically with respect to MedErr, our approach achieved $10.4$, which is lower than the prior state-of-the-art result reported in Mousavian et al~\cite{mousavian20173d}. 
In terms of $Acc_{\frac{\pi}{6}}$, our method outperforms the state-of-the-art result of Su et al~\cite{su2015render}. 
This is a quite positive result, since \cite{su2015render} uses additional rendered images for training.

We further evaluate $\mathit{Acc}_\frac{\pi}{18}$, which assesses the percentage of very accurate predictions. In this case, we simply compare against our re-implemented Res18, which achieved similar results with other state-of-the-art techniques. 
As shown in Table~\ref{tab:pascal_viewpoint}, our approach is significantly better than Res18-General/Specific with respect to $\mathit{Acc}_\frac{\pi}{18}$. This shows the advantage of performing keypoint alignment for pose estimation. 

Note that it is also possible to directly align CanViewFeature with StarMap for viewpoint estimation by a weak-perspective PnP~\cite{pavlakos20176} algorithm (\textbf{PnP} in Table~\ref{tab:pascal_viewpoint}).
In this case, utilizing DepthMap outperforms the direct alignment by $8.1\%$ in terms of $\mathit{Acc}_\frac{\pi}{6}$ and $1.75\%$ in terms of $\mathit{Acc}_\frac{\pi}{18}$, respecctively. 
On one hand, this shows the usefulness of DepthMap, particularly when the prediction is noisy. On the other hand, the performance of both approaches becomes similar when the predictions are very accurate ($\mathit{Acc}_\frac{\pi}{18}$). This is expected since both approaches should output identical results when the predictions are perfect.

\subsection{Analysis of Our Hybrid Keypoint Representation}
\label{Subsection:Advanced:Analysis}
\begin{table}[t]
\center
\scalebox{0.85}{
\begin{tabular}{l|cccccccccccc|c}
\hline
 & aero & bike & boat & bttl & bus & car  & chair & table & mbike & sofa & train & tv & mean \\
\hline
SIFT~\cite{long2014convnets} & 35 & 54 & 41 & 76 & 68 & 47 & 39 & 69 & 49 & 52 & 74 & 78 & 57\\
Conv~\cite{long2014convnets} & 44 & 53 & 42 & 78 & 70 & 45 & 41 & 68 & 53 & 52 & 73 & 76 & 58\\
Ours & 77 & 79 & 64 & 96 & 95 & 92 & 84 & 66 & 71 & 90 & 65 & 94 & 81 \\
\hline
\end{tabular}
}
\caption{Results for keypoint classification  on Pascal3D+ Dataset~\cite{xiang2014beyond}. We show keypoint classification accuracy of each category.}
\label{tab:kptclassification}
\end{table}

\noindent\textbf{Analysis of CanViewFeature.} 
We use the ground-truth keypoint location, and compare their learned 3D locations for keypoint classification with popular point features used in the literature, namely, SIFT~\cite{lowe2004distinctive} and Conv5 
of VGG~\cite{simonyan2014very}.
For CanViewFeature, we still follow the same procedure of using nearest neighbor for keypoint classification. 
For SIFT and Conv5, a linear SVM is used to classify the keypoints~\cite{long2014convnets}. 

Table~\ref{tab:kptclassification} compares CanViewFeature with the two baseline approaches from~\cite{long2014convnets}. We can see that CanViewFeature is significantly better than baseline approaches. This shows the advantage of using a shared keypoint representation for training a general purpose keypoint detector. 

\begin{table*}[t]
\scriptsize
\centering
\resizebox{.9\columnwidth}{!}{%
\begin{tabular}{c | c c c c c c c c c c c c| c }
\hline
& aero &bike& boat& bottle& bus& car& chair& table& mbike& sofa& train& tv & mean\\
\hline
$\mathit{MedErr}$(Ours) &  10.1  & 14.5  & 30.0  & 9.1 & 3.1 & 6.5  & 11.0  &  23.7 & 14.1  & 11.1  & 7.4 & 13.0  & 10.43\\
$\mathit{MedErr}$(GT Star) &  9.2  & 13.3  & 31.3  & 8.2 & 3.1 & 5.7  & 10.7  & 78.2 & 13.8  & 10.1  & 7.0 & 13.4  & 9.92\\
$\mathit{MedErr}$(GT Star+SCSF) &  7.7  & 12.9  & 22.0  & 8.0 & 3.0 & 5.9 & 9.3 & 14.6 & 10.8  & 8.3  & 6.3 & 12.9  & 9.1\\
$\mathit{MedErr}$(GT Star+Depth) &  6.2  & 6.2  & 14.1  & 2.4 & 2.1 & 3.9 & 6.5 & 72.9 & 7.0  & 5.4  & 6.8 & 1.9  & 4.7\\
\hline
$\mathit{Acc}_\frac{\pi}{6}$(Ours)& 0.82 & 0.86 & 0.50 & 0.92 & 0.97 & 0.92 & 0.79 & 0.62 & 0.88 & 0.92 & 0.77 & 0.83 & 0.8225  \\
$\mathit{Acc}_\frac{\pi}{6}$(GT Star)& 0.85 & 0.84 & 0.50 & 0.92 & 0.96 & 0.93 & 0.80 & 0.38 & 0.85 & 0.90 & 0.77 & 0.82 & 0.8211  \\
$\mathit{Acc}_\frac{\pi}{6}$(GT Star+SCSF)& 0.86 & 0.84 & 0.63 & 0.95 & 0.99 & 0.95 & 0.88 & 0.62 & 0.84 & 0.92 & 0.88 & 0.85 & 0.8651  \\
$\mathit{Acc}_\frac{\pi}{6}$(GT Star+Depth)& 0.86 & 0.93 & 0.63 & 0.95 & 0.97 & 0.91 & 0.82 & 0.38 & 0.87 & 0.92 & 0.84 & 0.93 & 0.8637  \\
\hline
\end{tabular}
}
\caption{Error Analysis on Pascal3D+. We show results in Median Error and Accuracy.}
\label{tab:error_analysis}
\end{table*}

\begin{table*}
     \begin{center}
     \begin{tabular}{cccccc}
     Input & StarMap & LocalMax. & CanViewFeat & Pred. 3D & Viewpoint \\
     \includegraphics[width=0.13\textwidth]{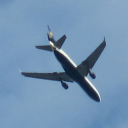}
      &\includegraphics[width=0.13\textwidth]{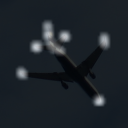}
      &\includegraphics[width=0.13\textwidth]{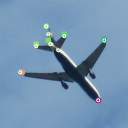}
      &\includegraphics[width=0.2\textwidth]{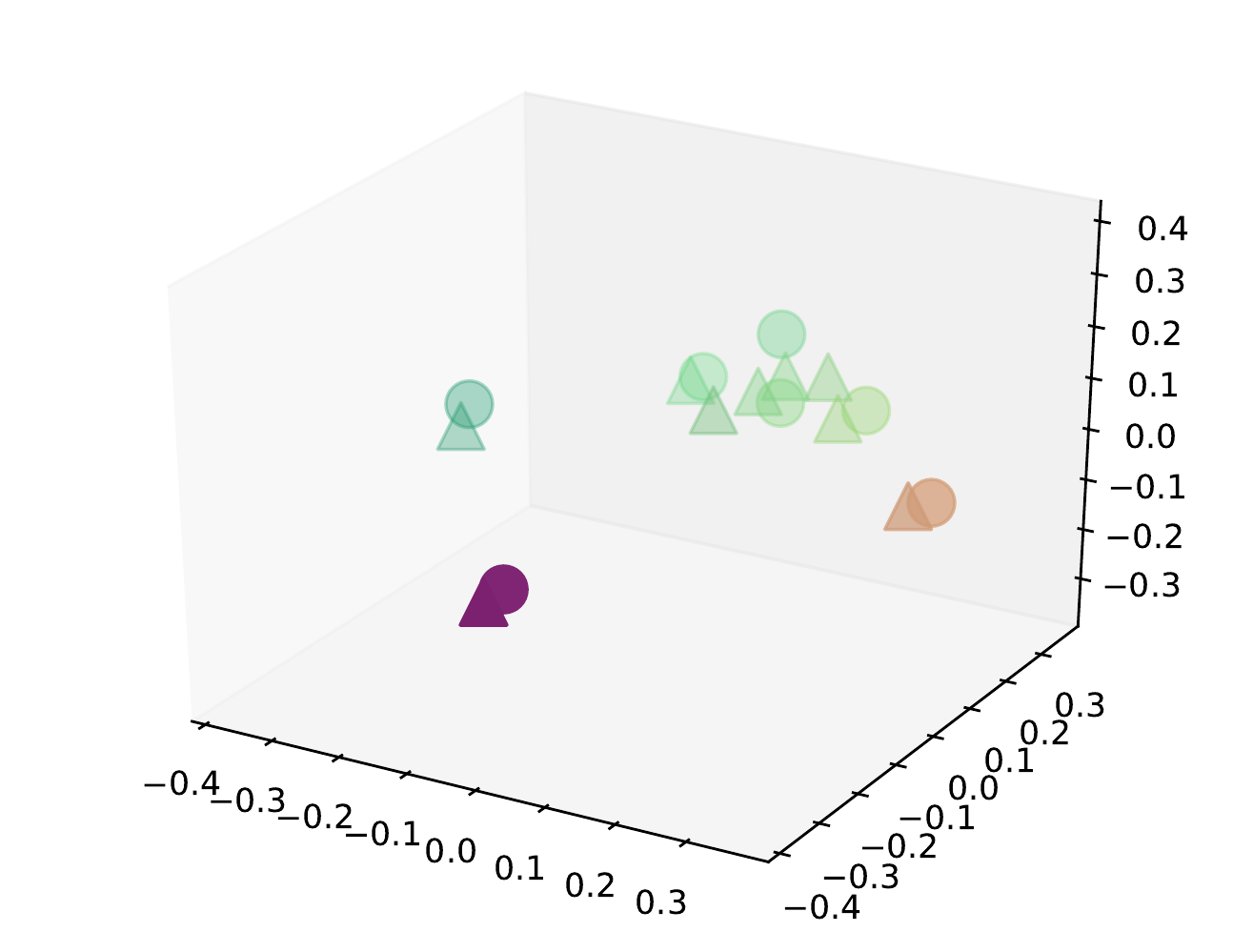}
      &\includegraphics[width=0.2\textwidth]{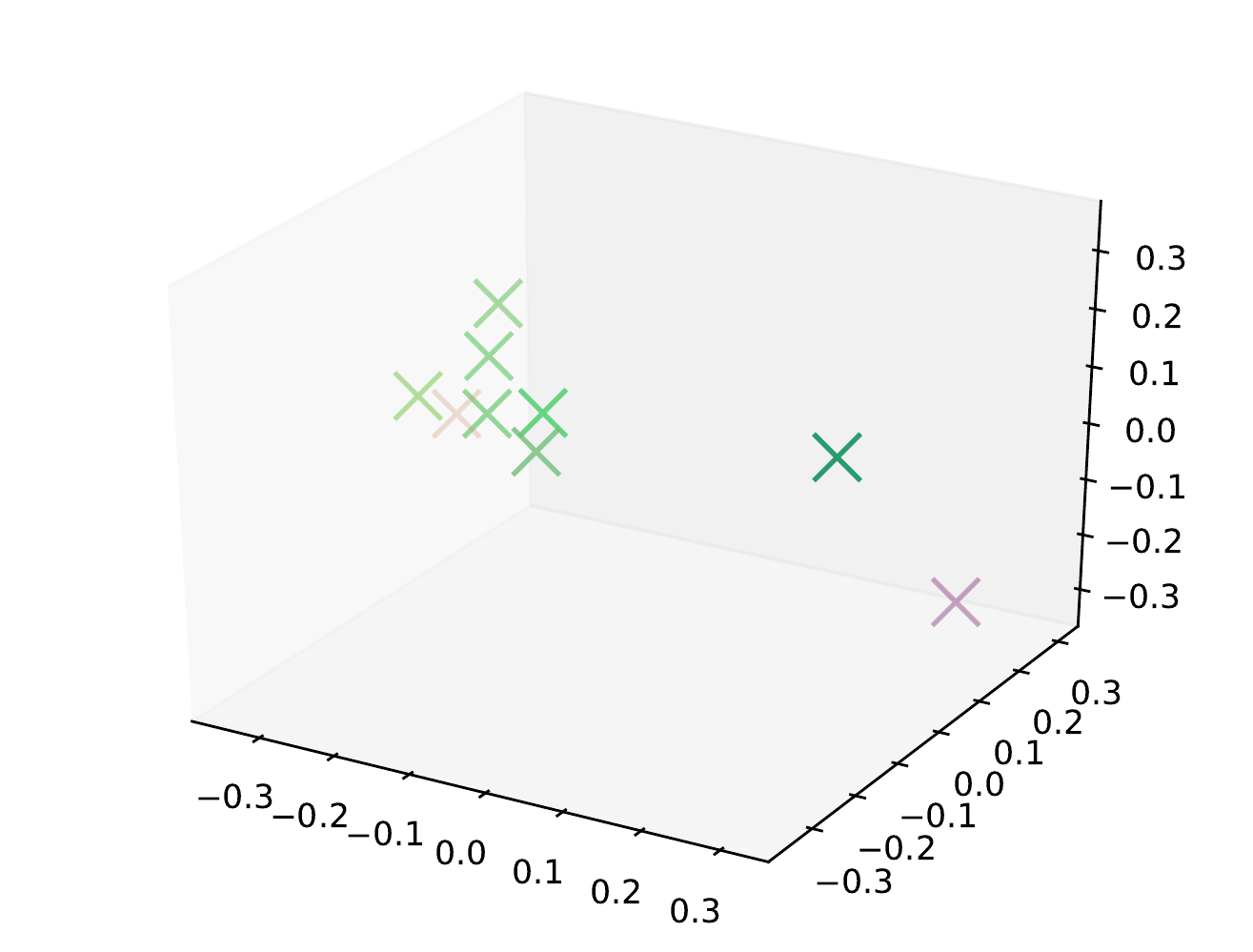}
      &\includegraphics[width=0.2\textwidth]{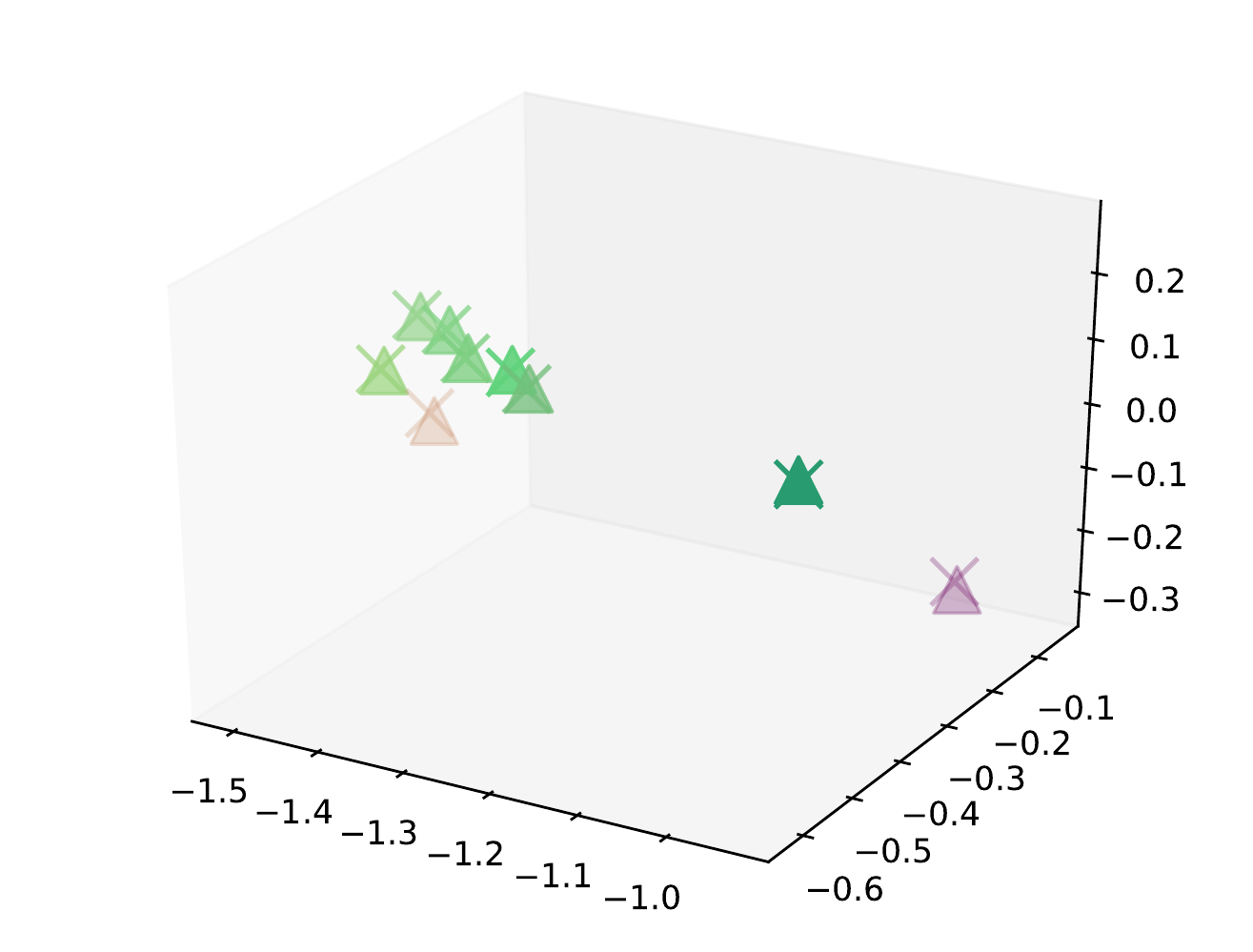} \\
     \includegraphics[width=0.13\textwidth]{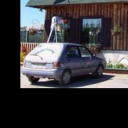}
      &\includegraphics[width=0.13\textwidth]{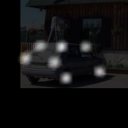}
      &\includegraphics[width=0.13\textwidth]{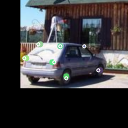}
      &\includegraphics[width=0.2\textwidth]{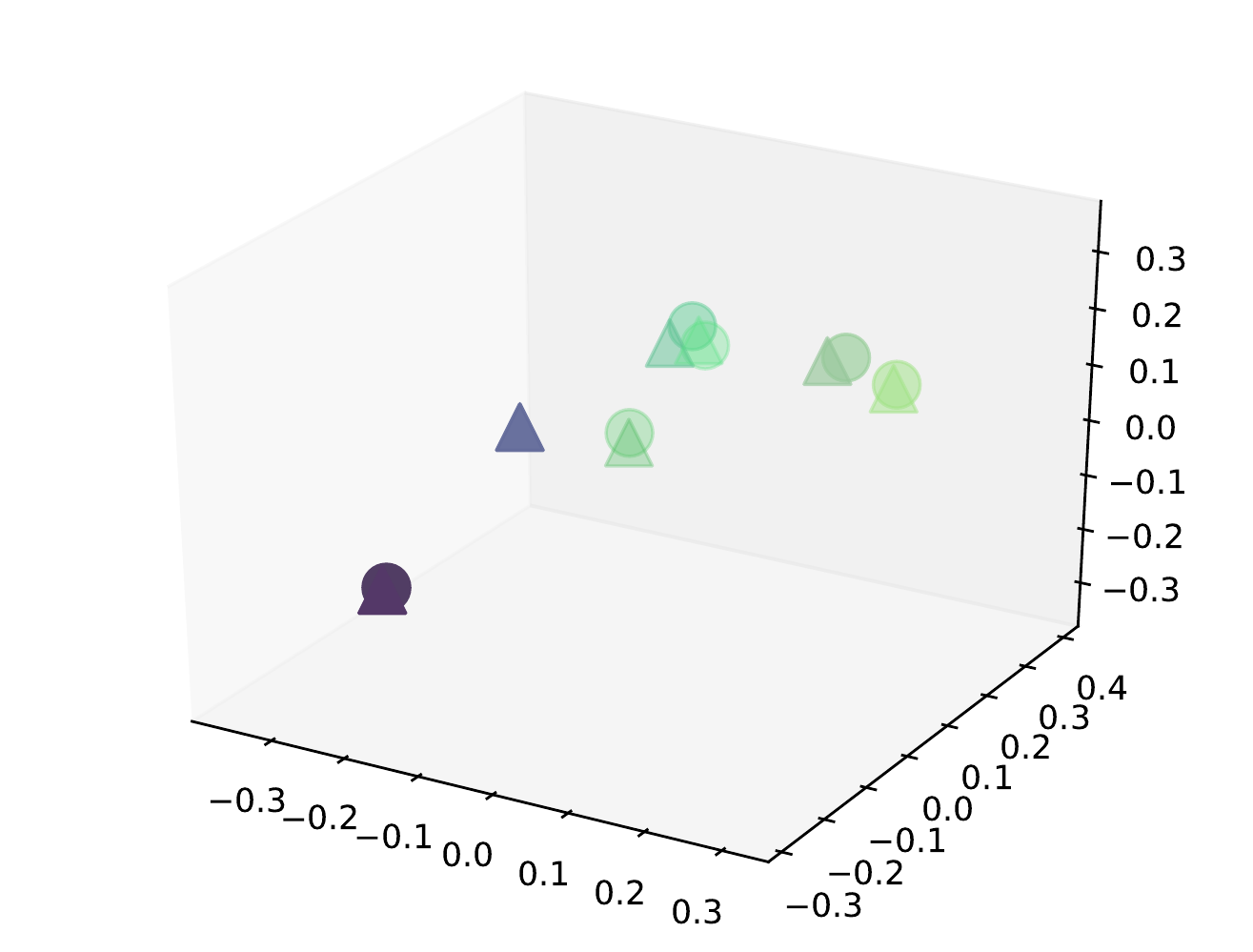}
      &\includegraphics[width=0.2\textwidth]{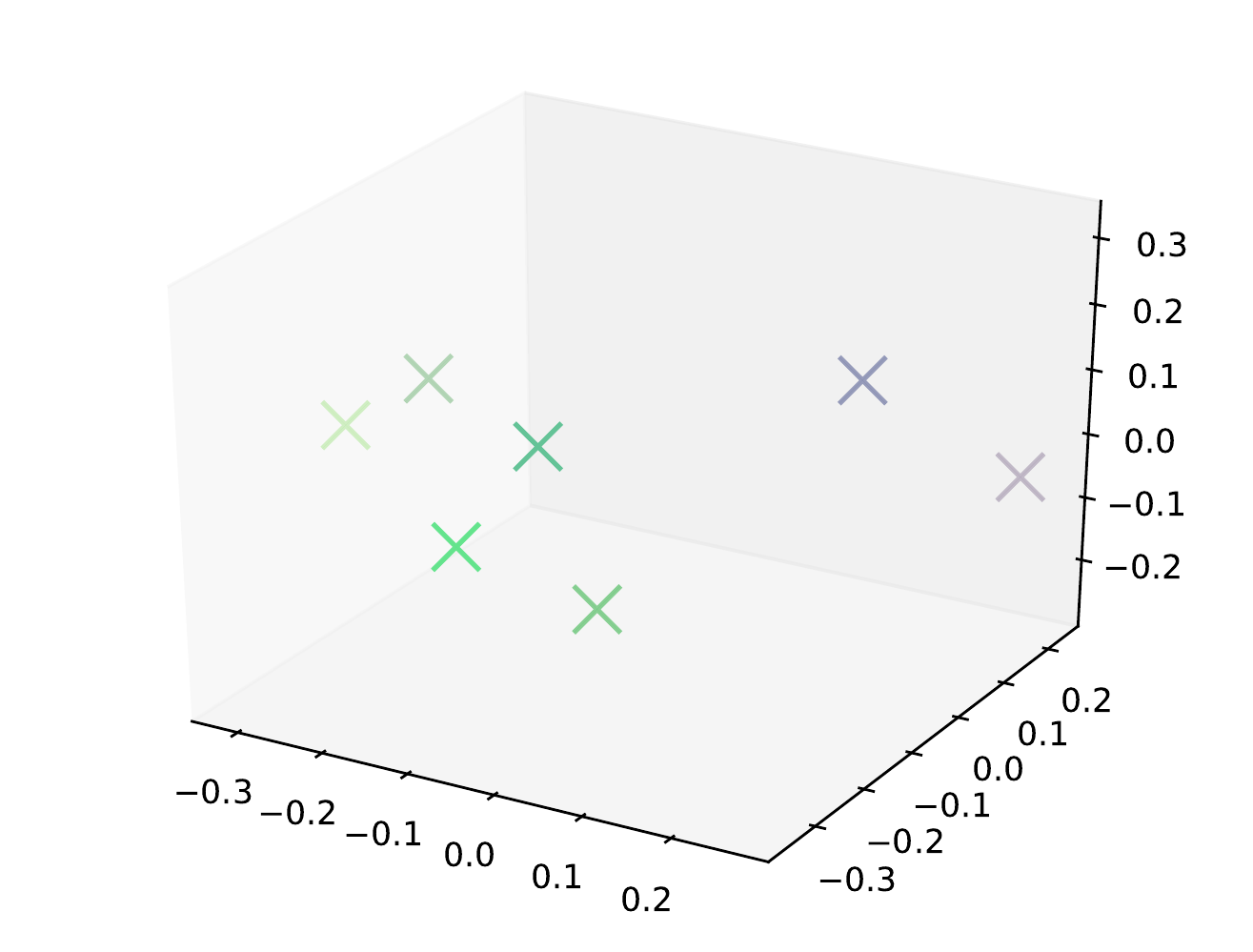}
      &\includegraphics[width=0.2\textwidth]{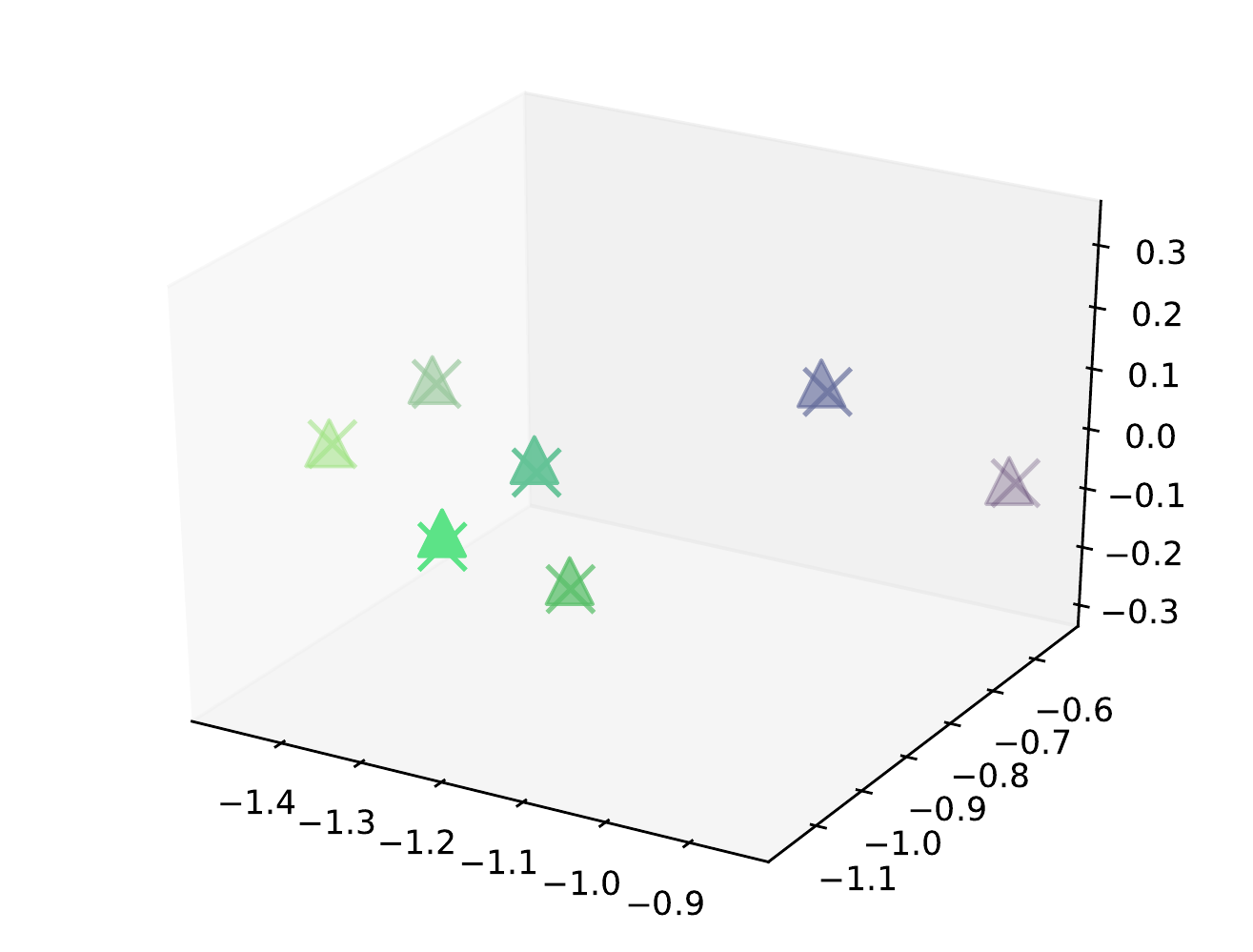} \\
     \includegraphics[width=0.13\textwidth]{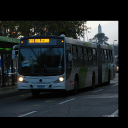}
      &\includegraphics[width=0.13\textwidth]{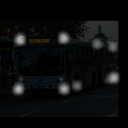}
      &\includegraphics[width=0.13\textwidth]{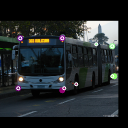}
      &\includegraphics[width=0.2\textwidth]{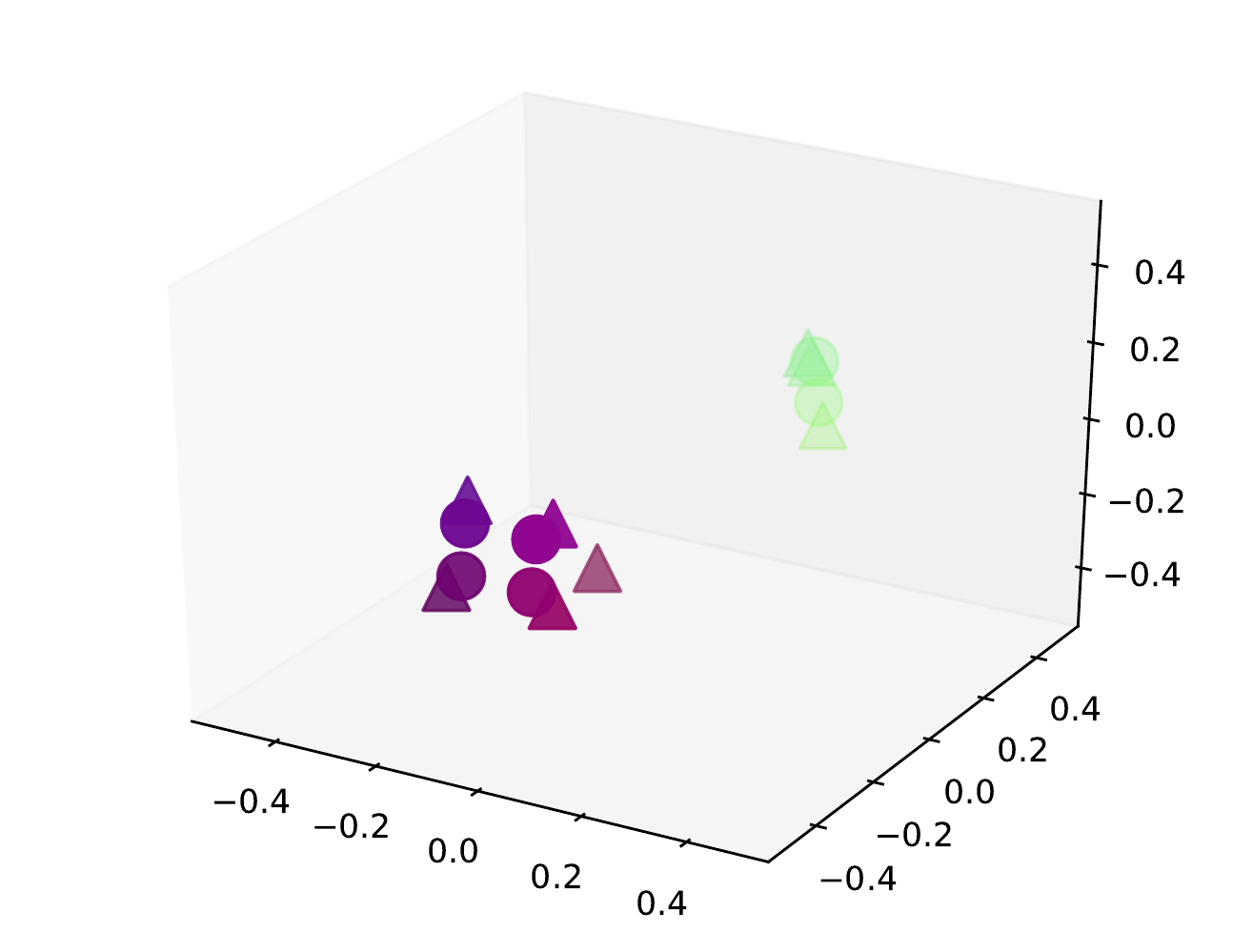}
      &\includegraphics[width=0.2\textwidth]{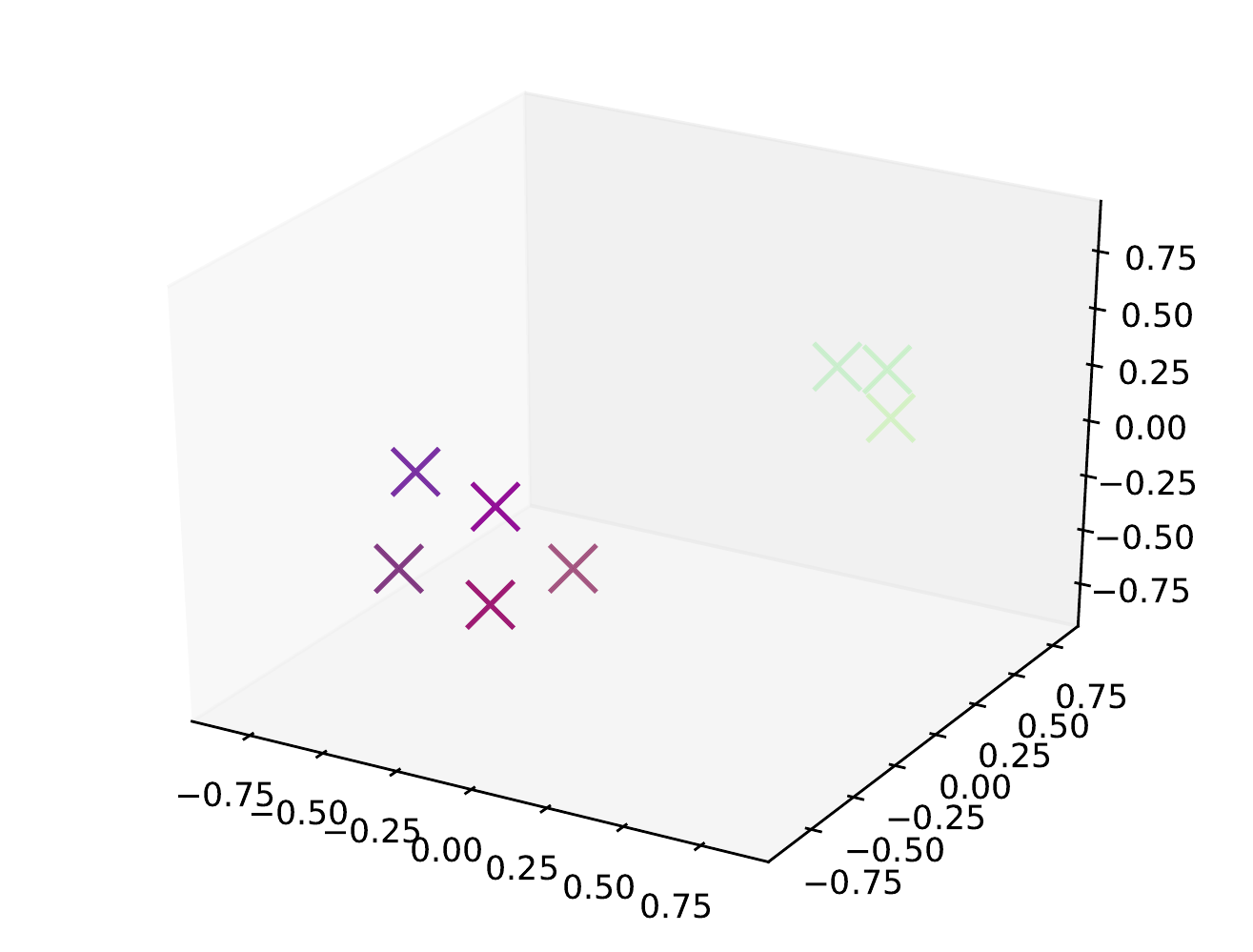}
      &\includegraphics[width=0.2\textwidth]{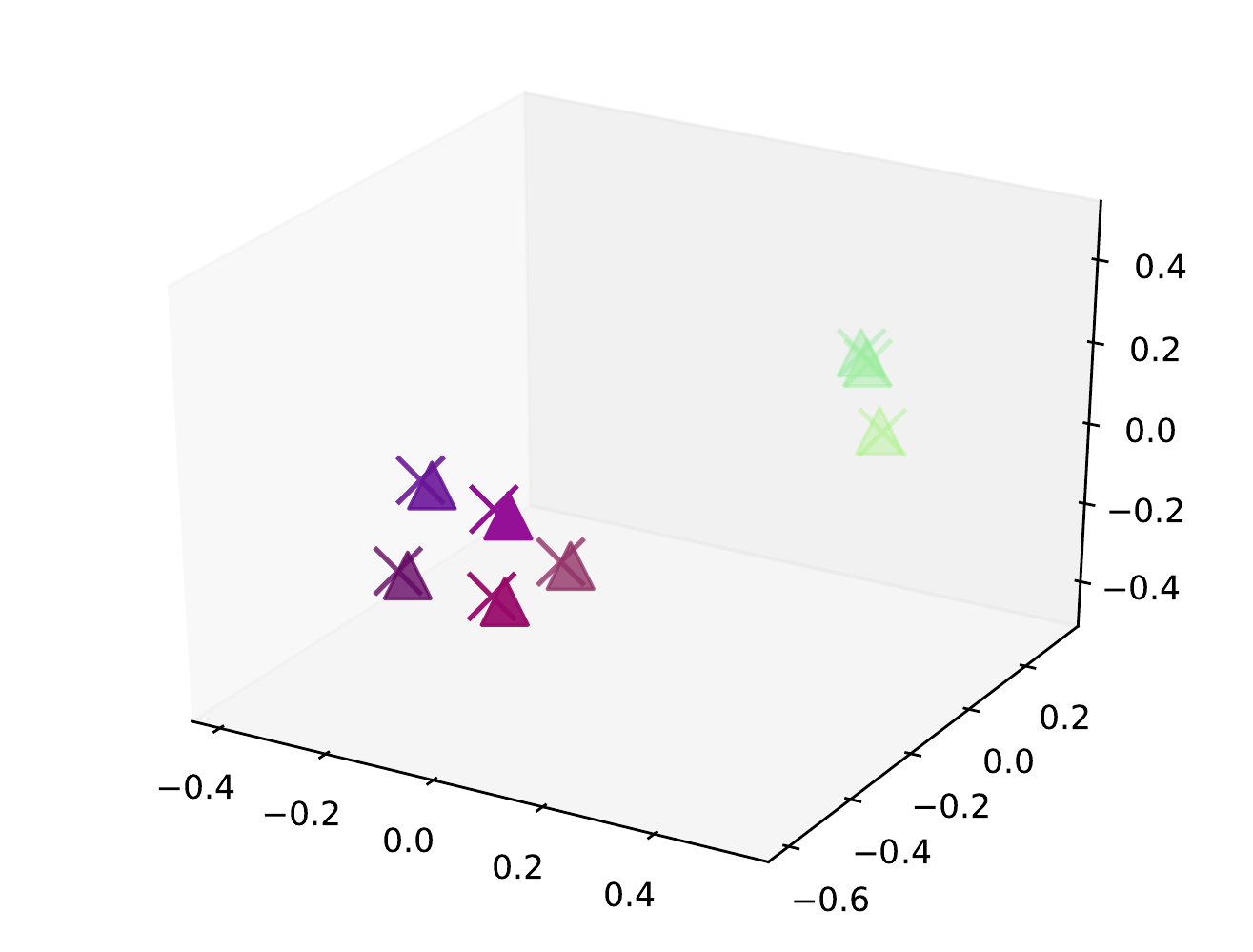} \\
     \includegraphics[width=0.13\textwidth]{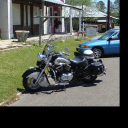}
      &\includegraphics[width=0.13\textwidth]{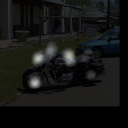}
      &\includegraphics[width=0.13\textwidth]{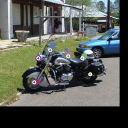}
      &\includegraphics[width=0.2\textwidth]{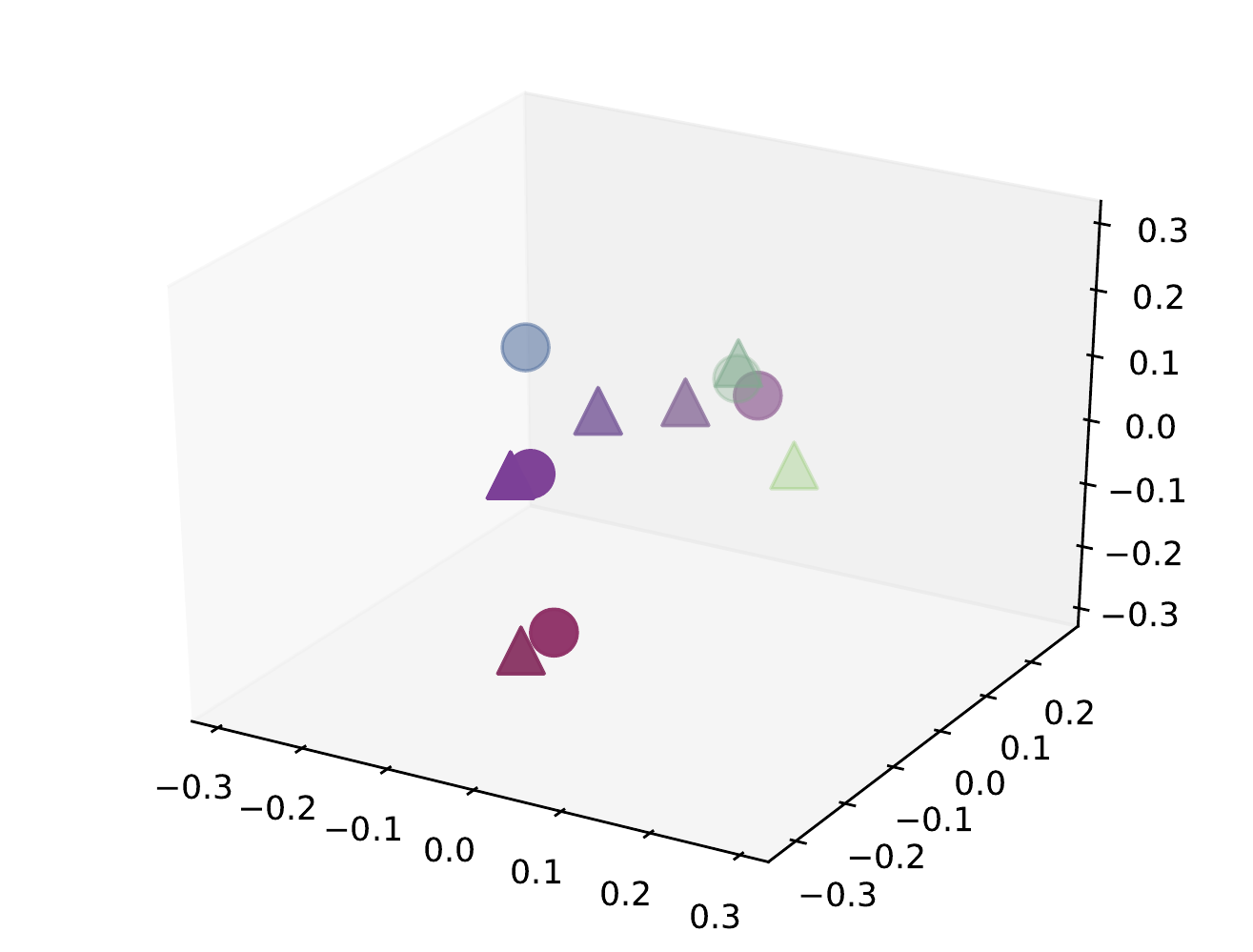}
      &\includegraphics[width=0.2\textwidth]{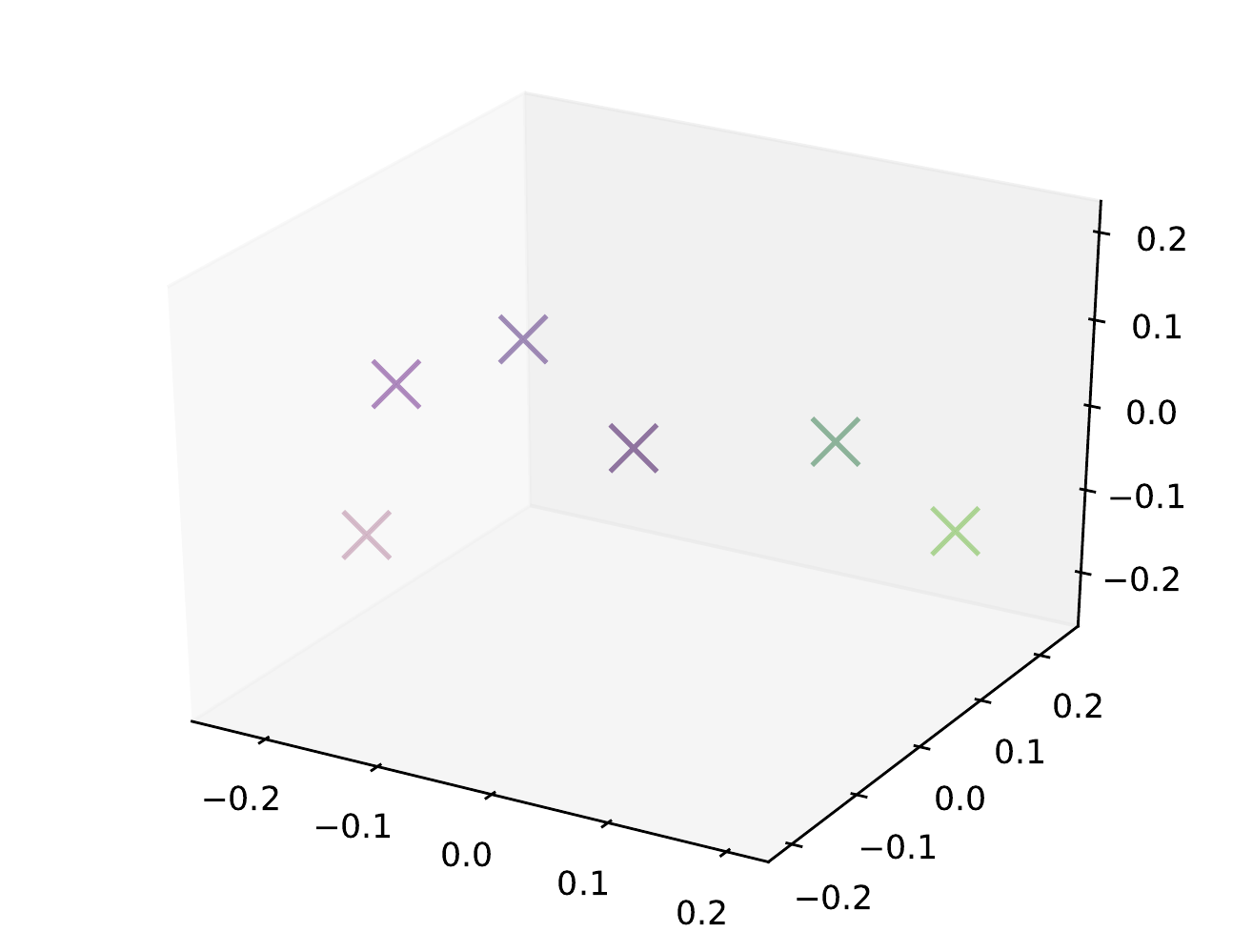}
      &\includegraphics[width=0.2\textwidth]{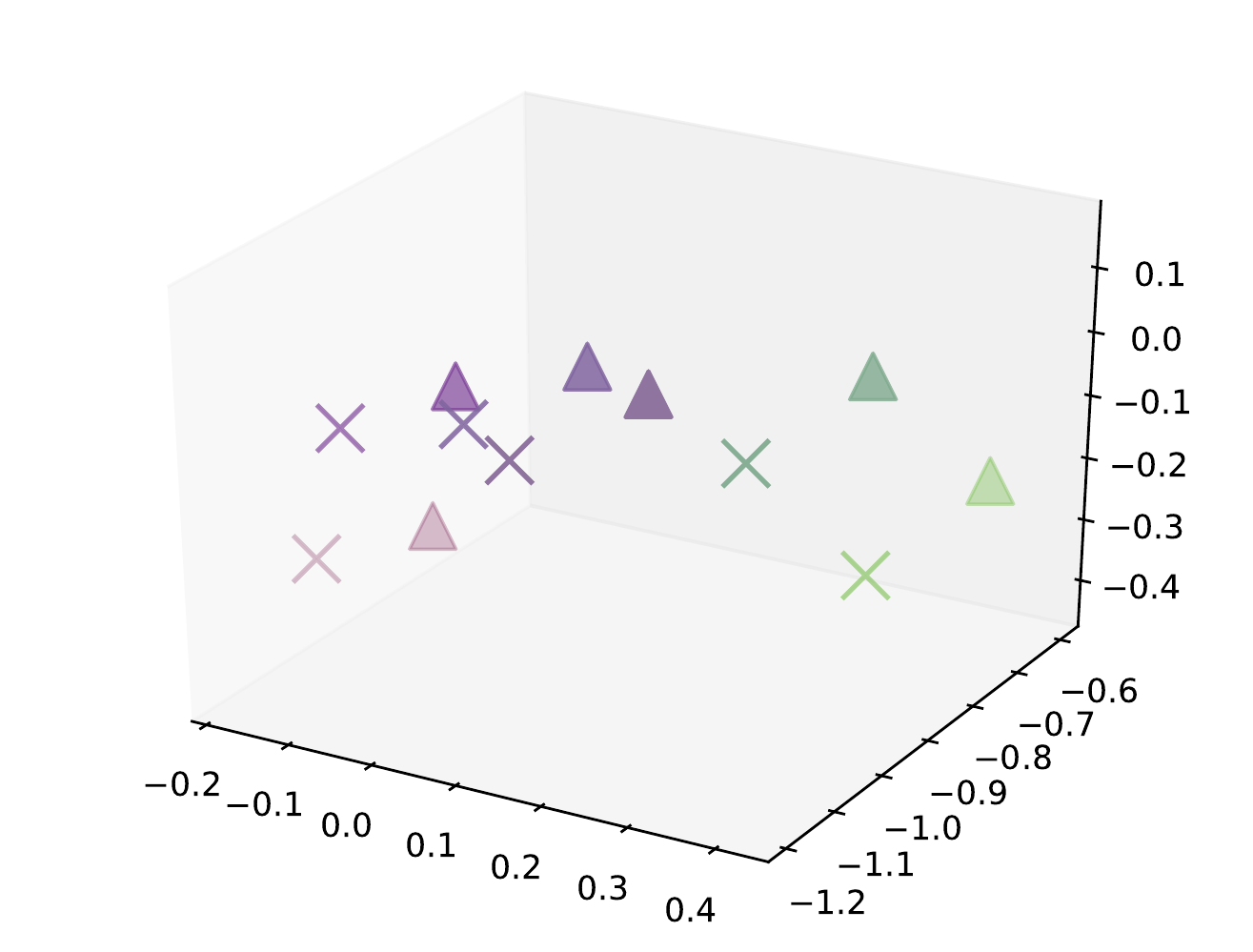} \\
     \includegraphics[width=0.13\textwidth]{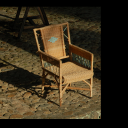}
      &\includegraphics[width=0.13\textwidth]{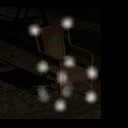}
      &\includegraphics[width=0.13\textwidth]{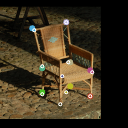}
      &\includegraphics[width=0.2\textwidth]{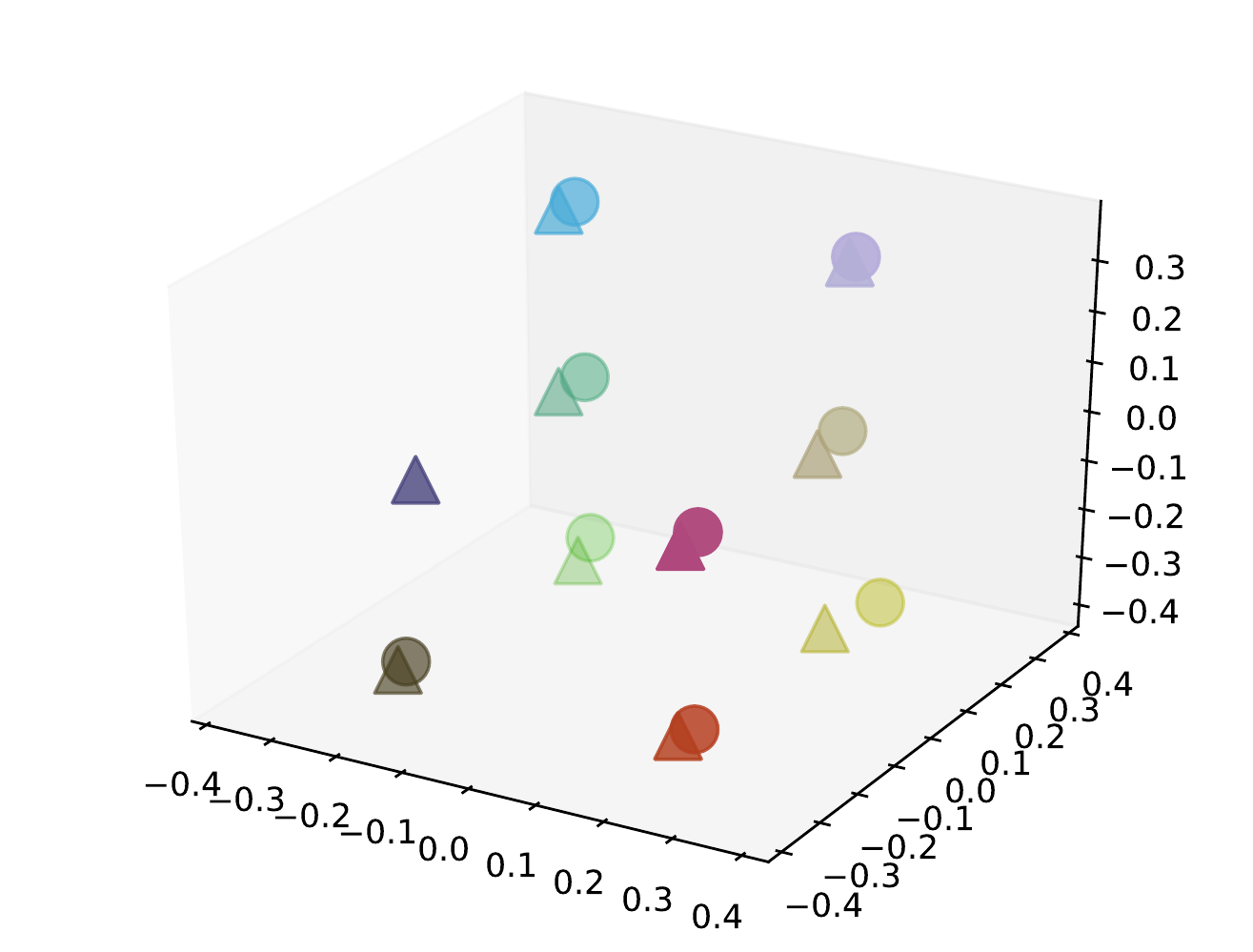}
      &\includegraphics[width=0.2\textwidth]{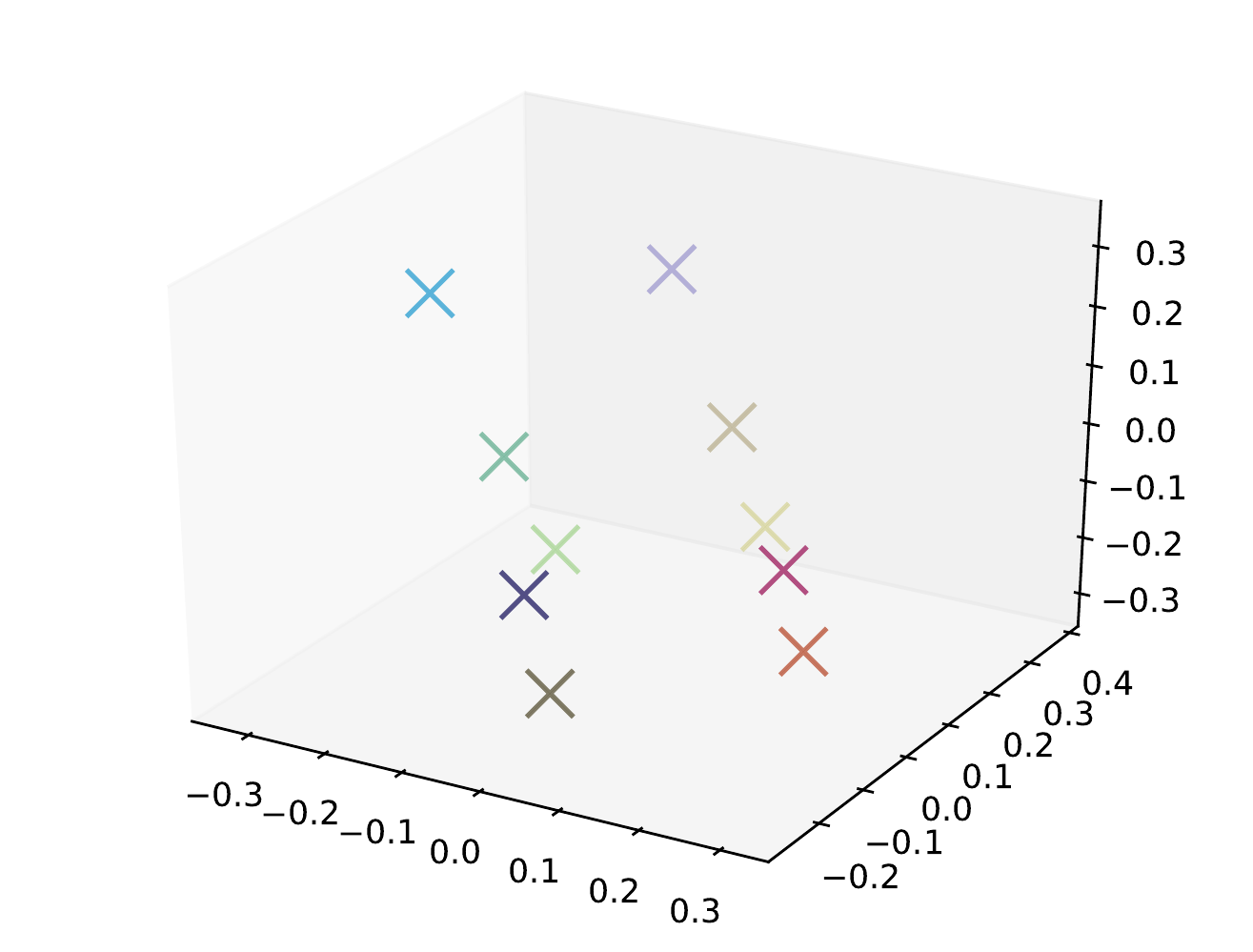}
      &\includegraphics[width=0.2\textwidth]{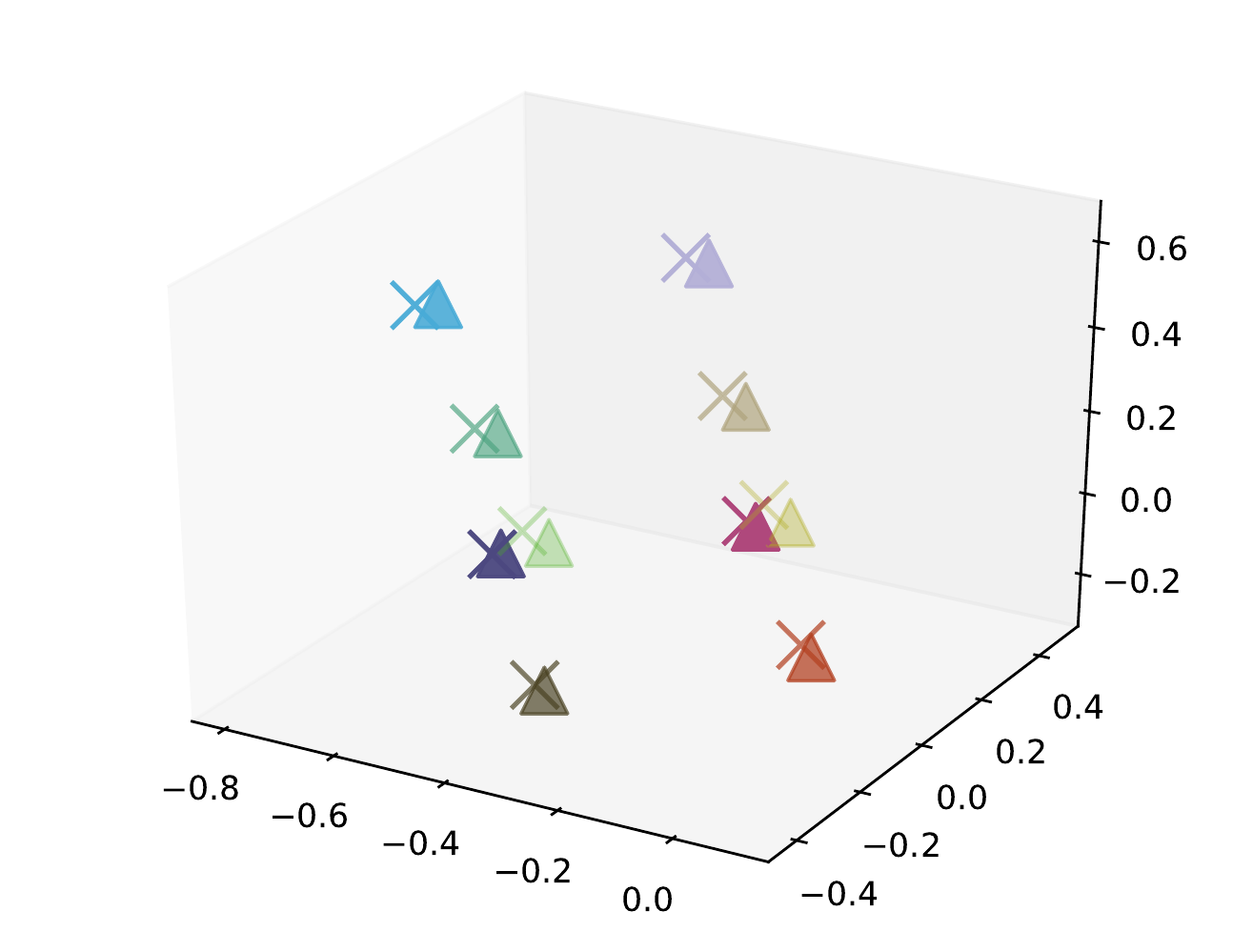} \\
     \includegraphics[width=0.13\textwidth]{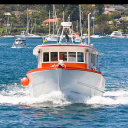}
      &\includegraphics[width=0.13\textwidth]{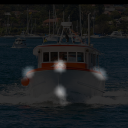}
      &\includegraphics[width=0.13\textwidth]{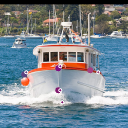}
      &\includegraphics[width=0.2\textwidth]{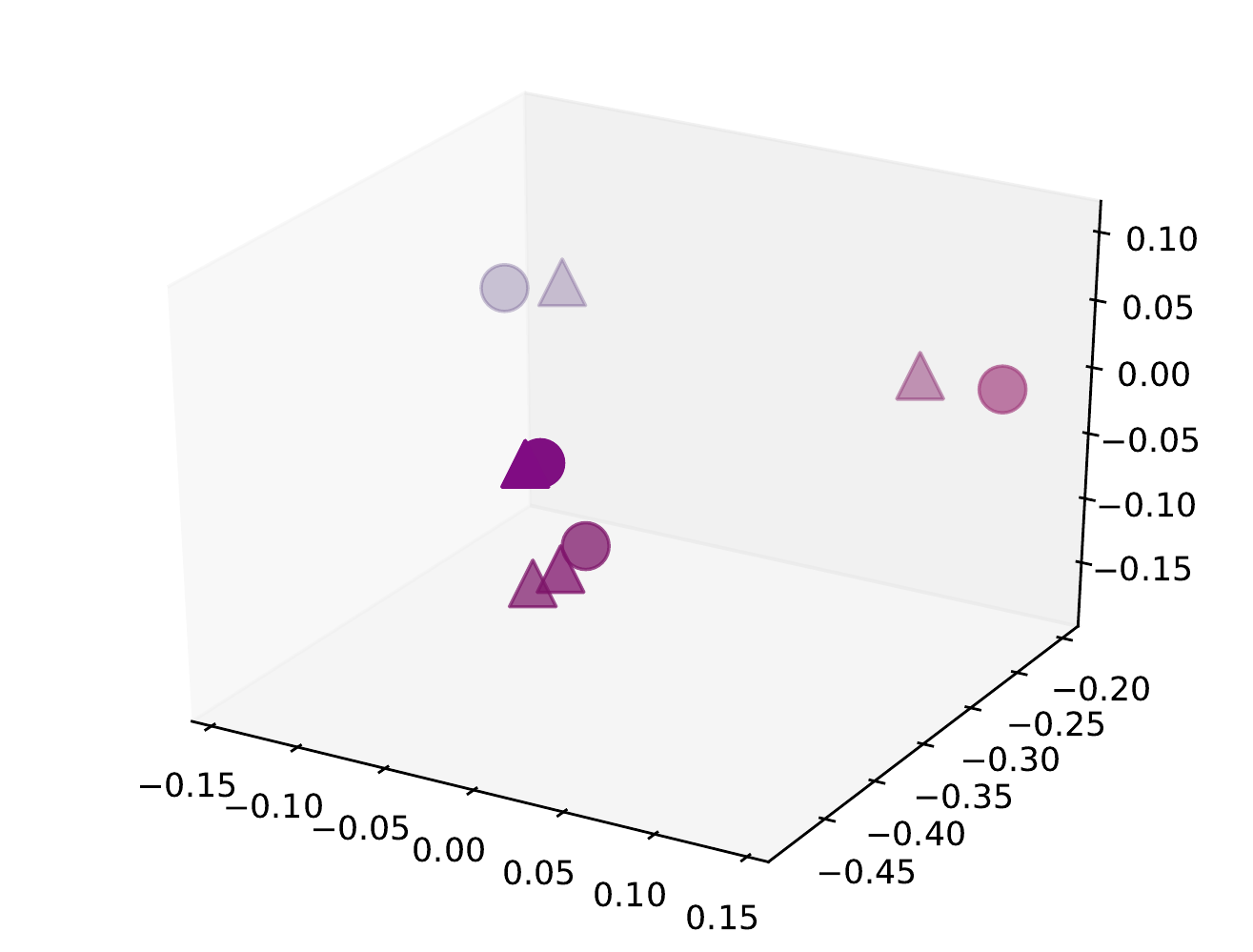}
      &\includegraphics[width=0.2\textwidth]{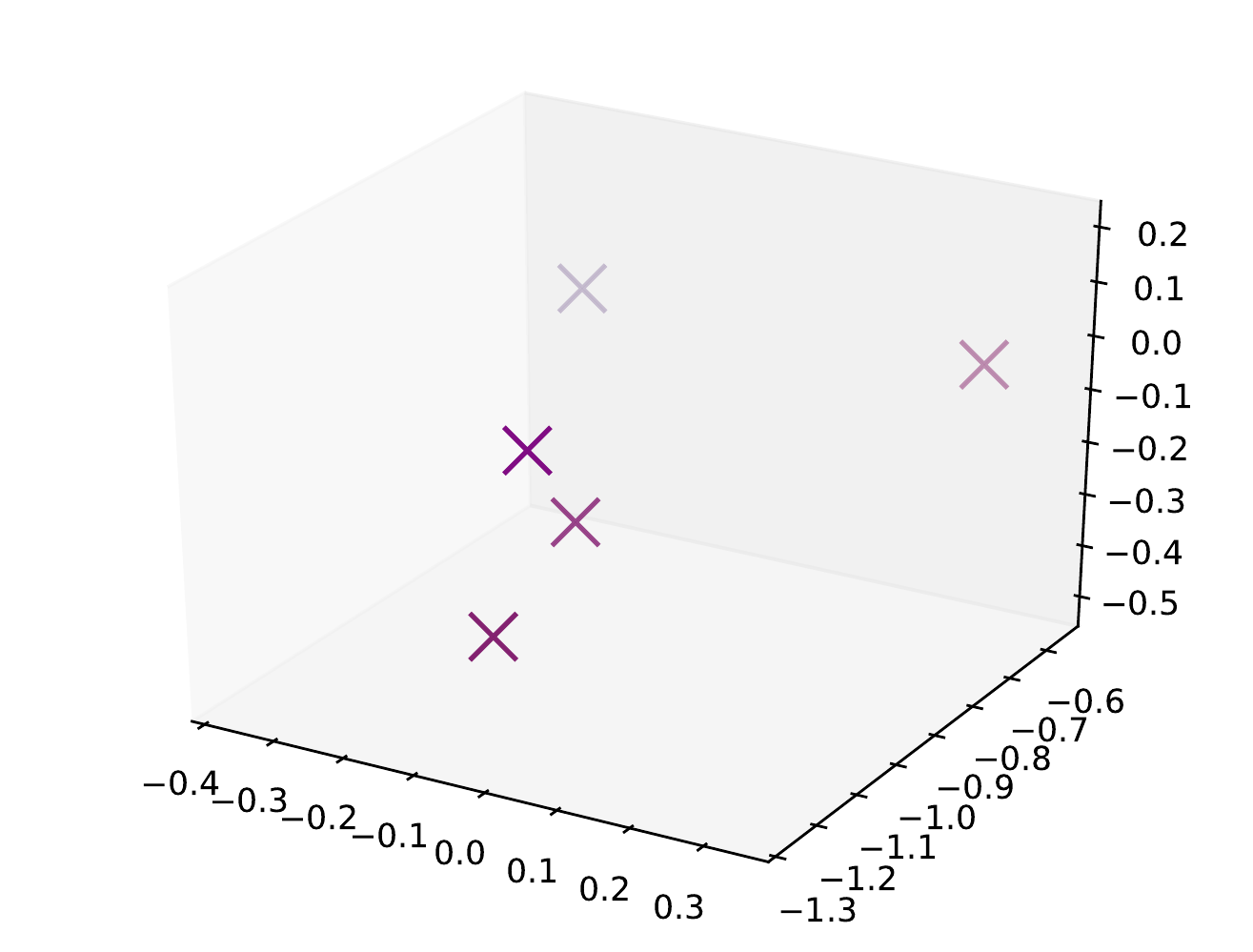}
      &\includegraphics[width=0.2\textwidth]{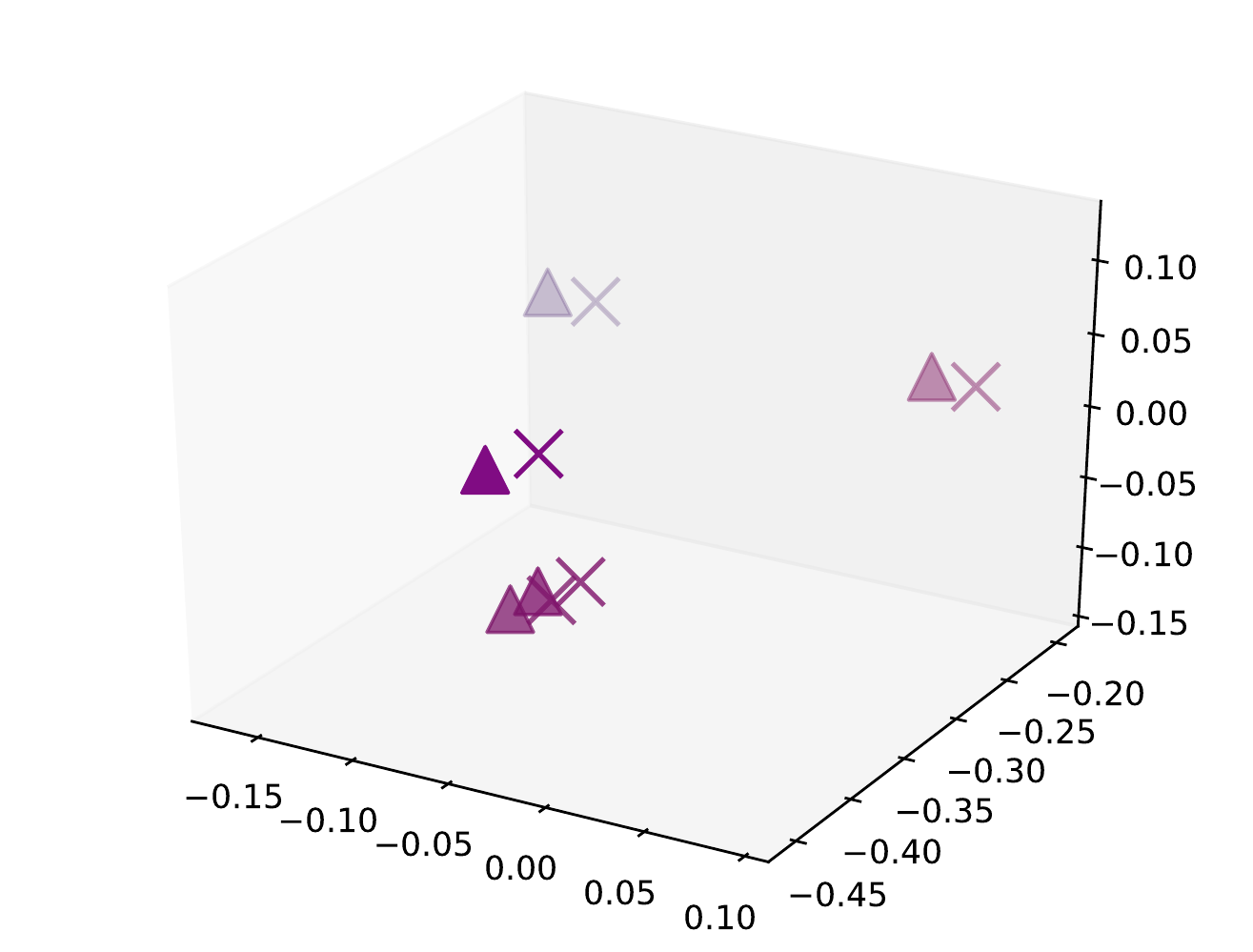} \\
     \includegraphics[width=0.13\textwidth]{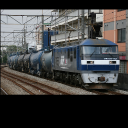}
      &\includegraphics[width=0.13\textwidth]{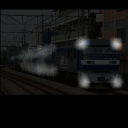}
      &\includegraphics[width=0.13\textwidth]{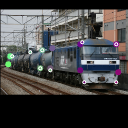}
      &\includegraphics[width=0.2\textwidth]{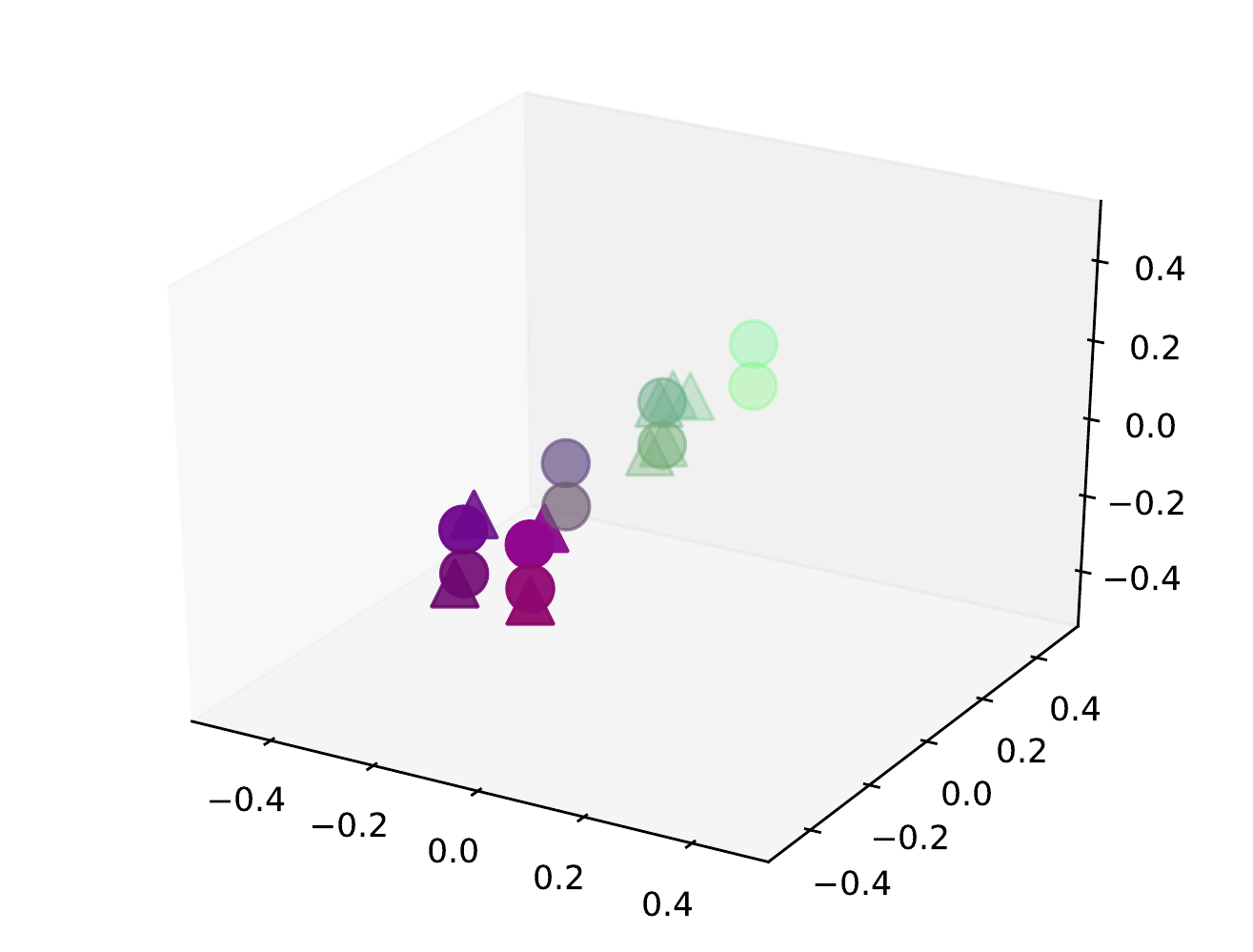}
      &\includegraphics[width=0.2\textwidth]{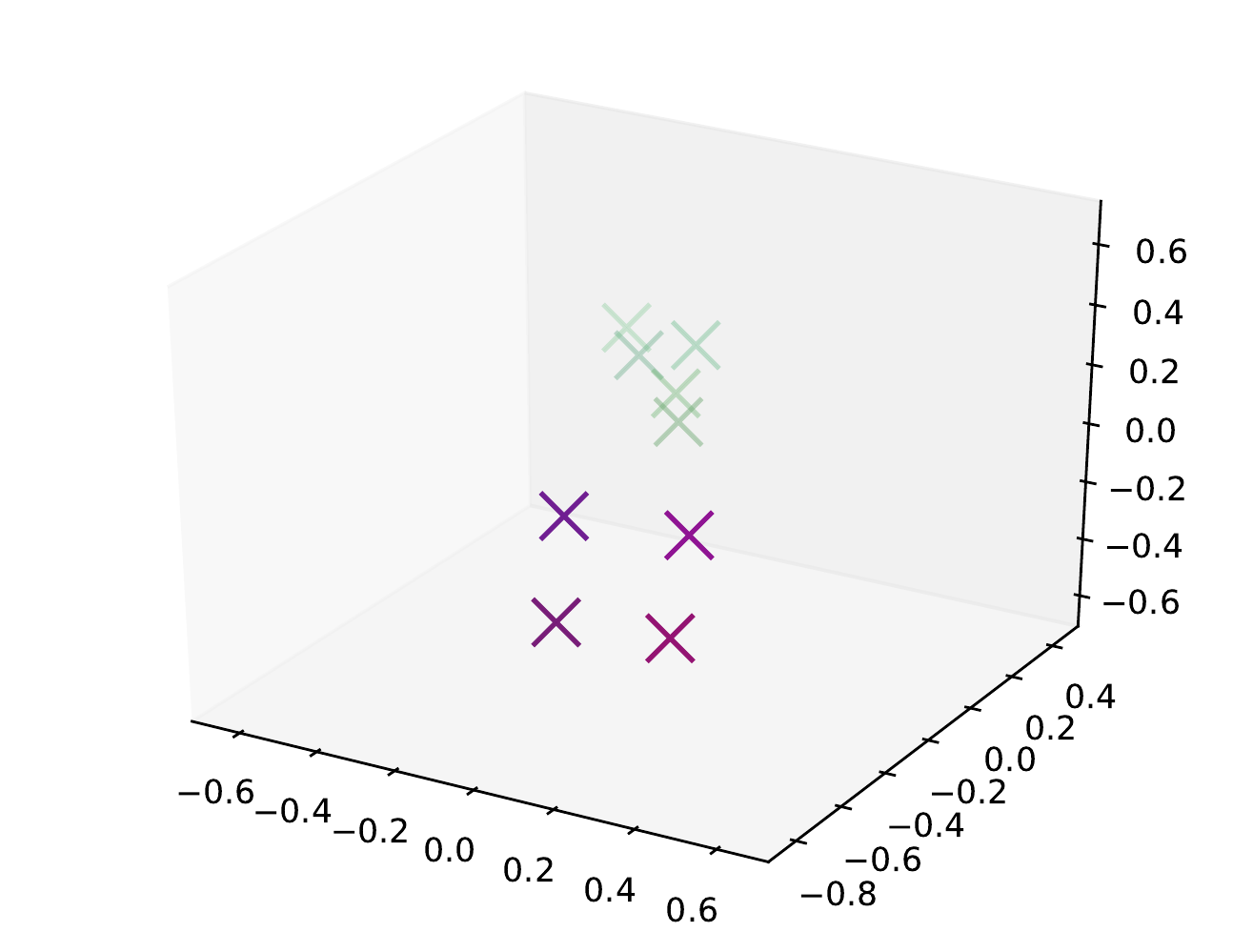}
      &\includegraphics[width=0.2\textwidth]{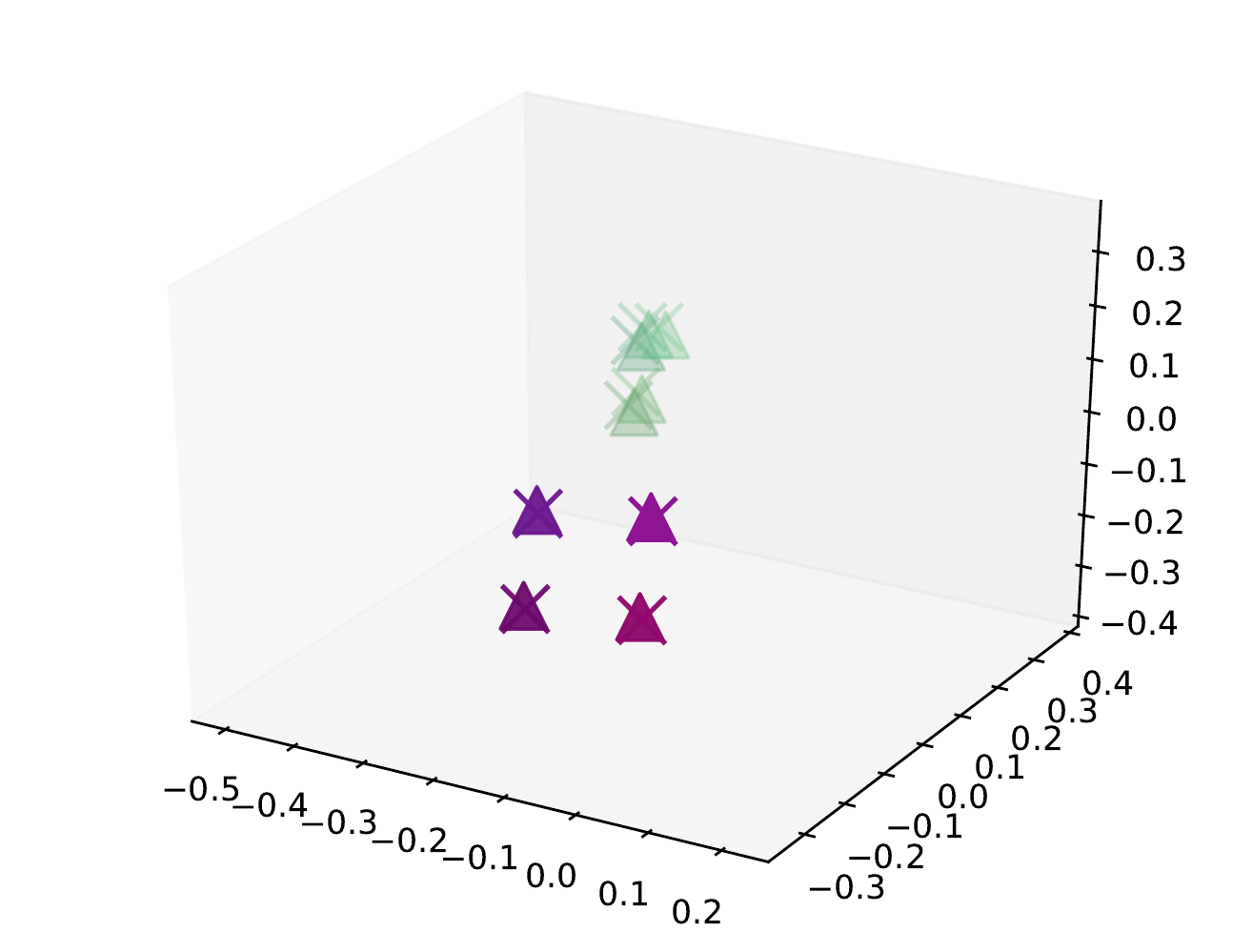} \\
     \includegraphics[width=0.13\textwidth]{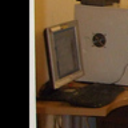}
      &\includegraphics[width=0.13\textwidth]{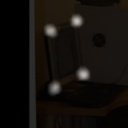}
      &\includegraphics[width=0.13\textwidth]{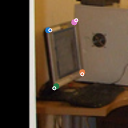}
      &\includegraphics[width=0.2\textwidth]{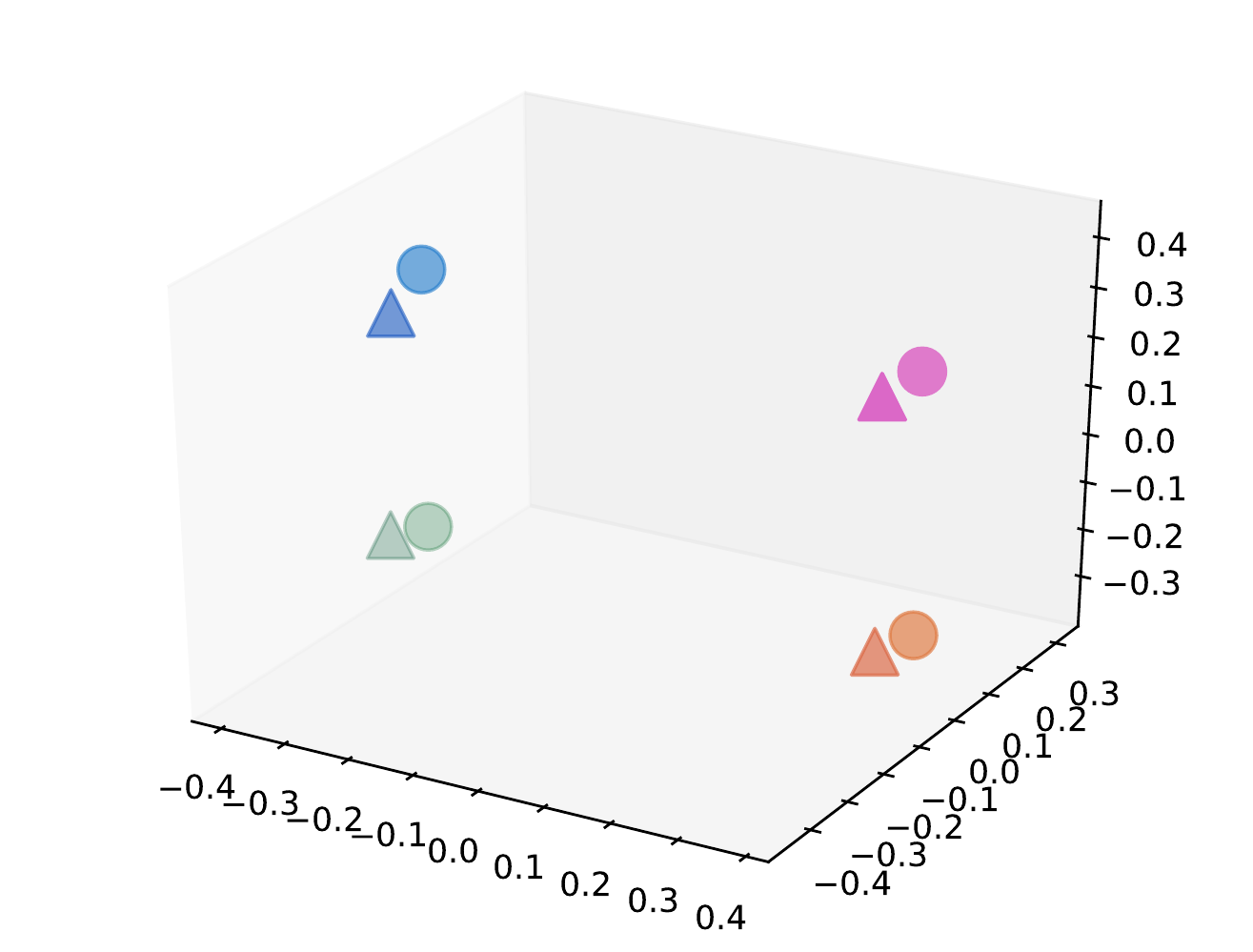}
      &\includegraphics[width=0.2\textwidth]{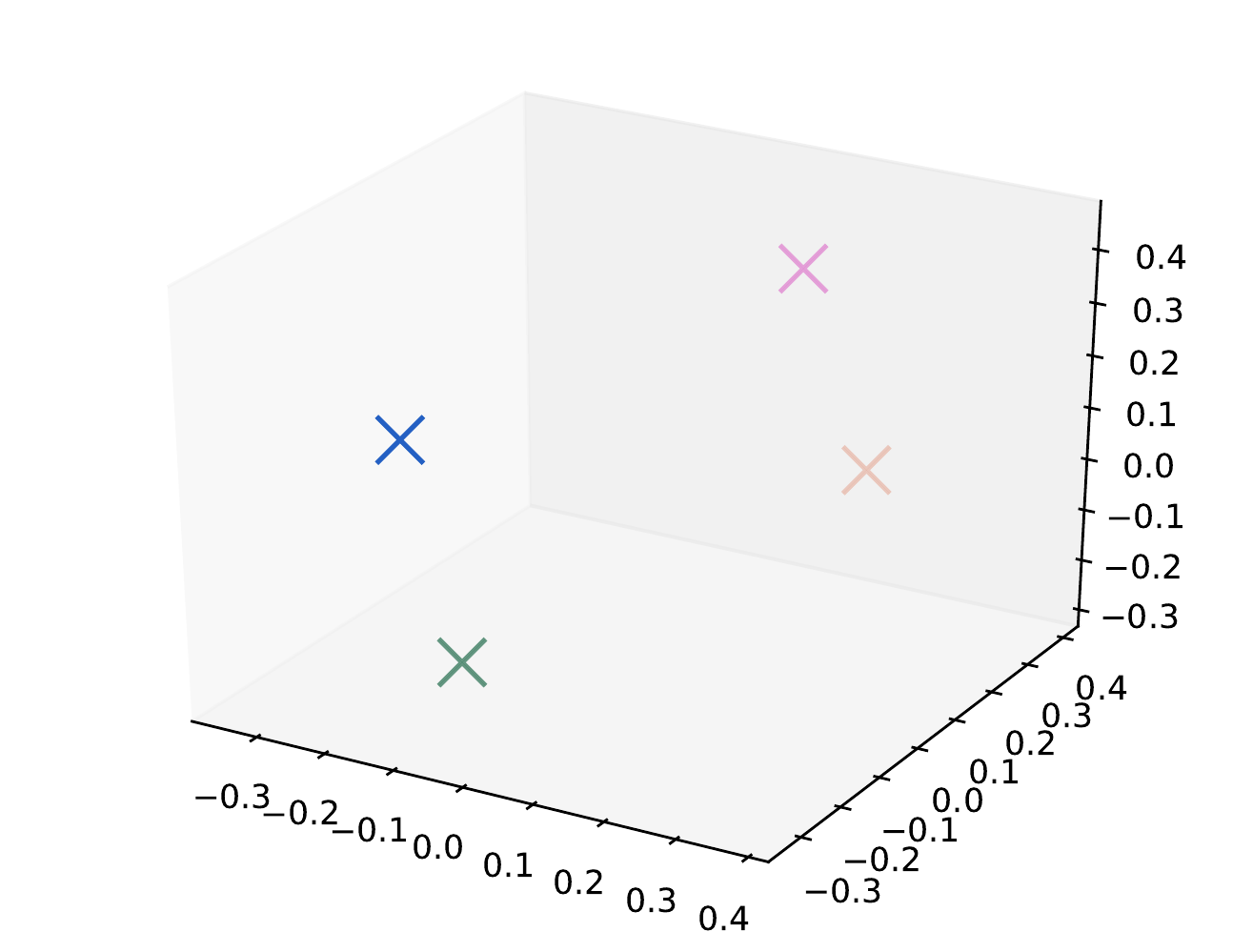}
      &\includegraphics[width=0.2\textwidth]{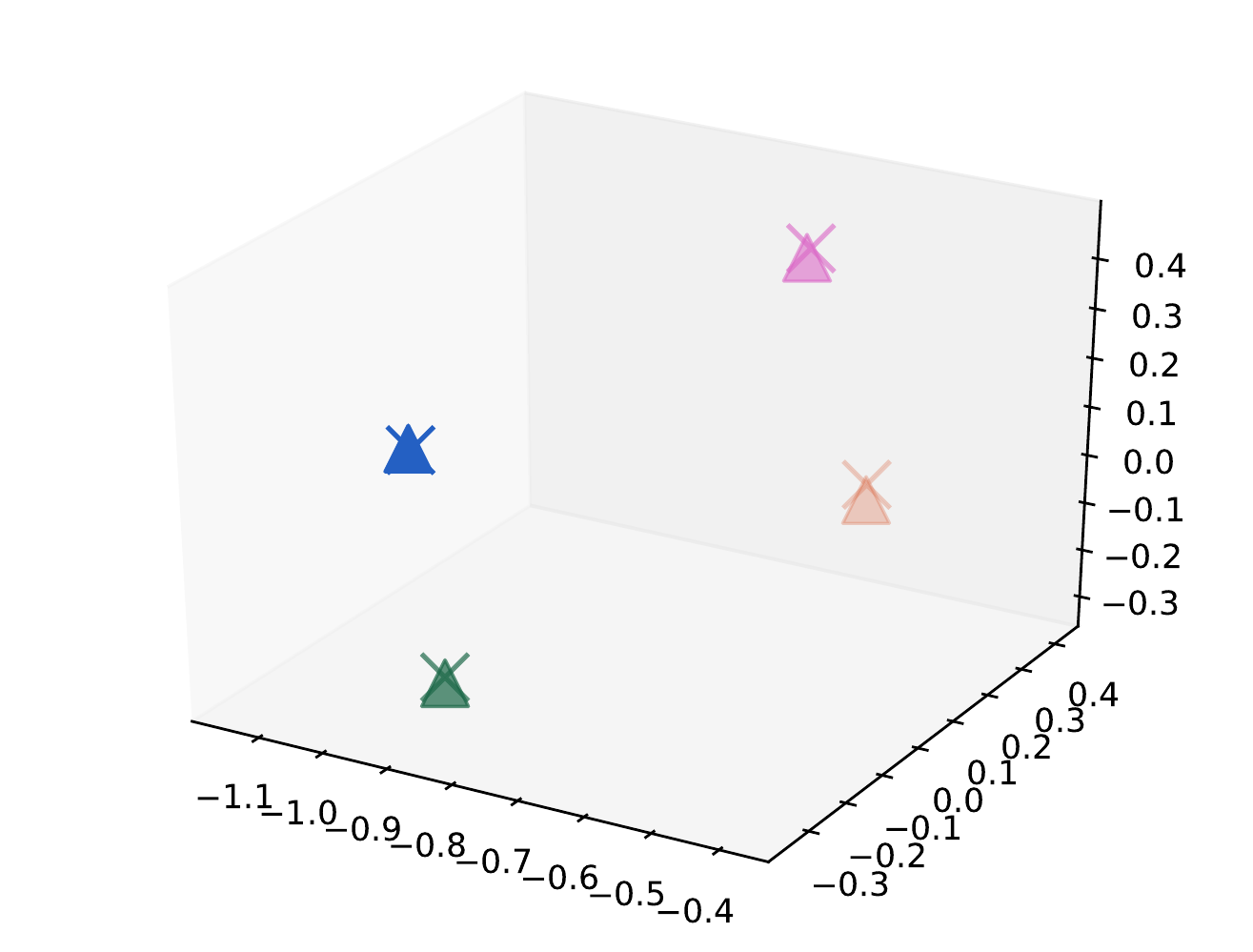} \\
      \end{tabular}
      \caption{Qualitative results of our full pipeline on Pascal3D+~\cite{xiang2014beyond} Dataset. 1st column: the input image; 2nd column: our predicted StarMap (shown on image); 3rd column: extracted keypoints after taking local maximum on StarMap, we show ground truth in large dots and prediction in small circled dots (The RGB color of the point encodes xyz coordinate for correspondence; 4th column: our predicted CanViewFeature (triangle) and their ground truth (circle); 5th column: our prediced 3D uvd coordinates, obtained by uv from StarMap and d from DepthMap; 6th column: rotated 3D point with our predicted viewpoint (cross) and ground truth viewpoint (triangle). }
      \label{table:demo}
      \end{center}
\end{table*}

\noindent\textbf{Ablation study on representation components.} To better understand the importance of each component of our representation and whether they are well-trained, we provide error analysis by replacing each output component with its ground truth. To this end, we use viewpoint estimation as the task for evaluation, and Table~\ref{tab:error_analysis} summarizes the results. 
Specifically, replacing {StarMap} with its ground truth does not provides much performance gains in both metrics, indicating that {StarMap} is fairly accurate. 
This is justified by the high keypoint accuracy reported in  Section~\ref{Subsection:Keypoint:Detection:Classification}. 
Moreover, replacing either {CanViewFeature} or {DepthMap} with the underlying ground truth provides considerable performance gains in terms of $\mathit{Acc}_\frac{\pi}{6}$. 
In particular, using perfect {DepthMap} leads noticeable decrease in median error. 
This is expected since the general task of estimating pixel depth remains quite challenging.

\subsection{Keypoint and Viewpoint Induction for Novel categories}
\label{Subsection:PoseInduction}

\begin{table}[t]
\scriptsize
\center{
\begin{tabular}{l|ccccccccc}
 \hline
& bed  & bookshelf  & calculator  & cellphone  & computer  & filing cabinet  & guitar  & iron  & knife  \\
 \hline
$\mathit{Acc}_\frac{\pi}{6}$ (Sup) & 0.73 &  0.78 &  0.91 &  0.57 &  0.82 &  0.84 &  0.73 &  0.03 &  0.18 \\
$\mathit{Acc}_\frac{\pi}{6}$ (Novel) & 0.37 &  0.69 &  0.19 &  0.52 &  0.73 &  0.78 &  0.61 &  0.02 &  0.09 \\
\hline
& microwave  & pen  & pot  & rifle  & slipper  & stove  & toilet  & tub  & wheelchair  \\
 \hline
$\mathit{Acc}_\frac{\pi}{6}$ (Sup) & 0.94 &  0.13 &  0.56 &  0.04 &  0.12 &  0.87 &  0.71 &  0.51 &  0.60  \\
$\mathit{Acc}_\frac{\pi}{6}$ (Novel) & 0.88 &  0.12 &  0.51 &  0.00 &  0.11 &  0.82 &  0.41 &  0.49 &  0.14  \\
 \hline
\end{tabular}
}
{\caption{Viewpoint estimation for novel categories results on ObjectNet3D+~\cite{xiang2016objectnet3d}. We shown our results in $\mathit{Acc}_\frac{\pi}{6}$. }
\label{table:objectNet3D}}
\end{table}

Our keypoint representation is category-agnostic and is free to be extended to novel object categories~\cite{tulsiani2015pose}.

We note that Pascal3D+~\cite{xiang2014beyond} only contains $12$ categories and it is hard to learn common inter-category information with such limited category samples. 
To further verify the generalization ability of our method, we used a newly published large scale 3D dataset, ObjectNet3D~\cite{xiang2016objectnet3d}.
ObjectNet3D~\cite{xiang2016objectnet3d} has the same annotations as Pascal3D+~\cite{xiang2014beyond} but with 100 categories.
We evenly hold out 20 categories (every 5 categories sorted in the alphabetical order) from the training data and only used them for testing. Because Shoe and Door do not have keypoint annotation, we remove them from the testing set, resulting in 18 novel categories. Please refer to the supplementary for details on dataset details.

We compare the performance gap between including and withholding the $18$ categories during training. 
The results are shown in Table~\ref{table:objectNet3D}. 
As expected, the viewpoint estimation accuracy of most categories drops.
For some categories (Iron, Knife, Pen, Rifle, Slipper), both experiments fail (with accuracy lower than $20\%$). One explanation is that these 5 failed categories are small and narrow objects, whose annotations may not be accurate. For example, the keypoint annotations on ObjectNet3D~\cite{xiang2016objectnet3d} for small object are not always well-defined (see qualitative results in supplementary),
e.g., Key and Spoon have dense keypoints annotation on their silhouette.
For half of the $18$ novel objects (bookshelf, cellphone, computer, filing cabinet, guitar, microwave, pot, stove, tub), the performance gap between including and withholding training data is less than $10\%$.
This indicates that our representation is fairly general and can extend viewpoint estimation to novel categories.

\noindent\textbf{Acknowledgement.} We thank Shubham Tulsiani and Angela Lin for the helpful discussions. 

\bibliographystyle{splncs04}
\bibliography{egbib}

\begin{thebibliography}{10}
\providecommand{\url}[1]{\texttt{#1}}
\providecommand{\urlprefix}{URL }
\providecommand{\doi}[1]{https://doi.org/#1}

\bibitem{altwaijry2016learning}
Altwaijry, H., Veit, A., Belongie, S.J., Tech, C.: Learning to detect and match
  keypoints with deep architectures. In: BMVC (2016)

\bibitem{andriluka14cvpr}
Andriluka, M., Pishchulin, L., Gehler, P., Schiele, B.: 2d human pose
  estimation: New benchmark and state of the art analysis. In: IEEE Conference
  on Computer Vision and Pattern Recognition (CVPR) (June 2014)

\bibitem{bourdev2010detecting}
Bourdev, L., Maji, S., Brox, T., Malik, J.: Detecting people using mutually
  consistent poselet activations. In: European conference on computer vision.
  pp. 168--181. Springer (2010)

\bibitem{cao2017realtime}
Cao, Z., Simon, T., Wei, S.E., Sheikh, Y.: Realtime multi-person 2d pose
  estimation using part affinity fields. In: CVPR. vol.~1, p.~7 (2017)

\bibitem{LocalChapterEvents:ItalChap:ItalianChapConf2008:129-136}
Cignoni, P., Callieri, M., Corsini, M., Dellepiane, M., Ganovelli, F.,
  Ranzuglia, G.: {MeshLab: an Open-Source Mesh Processing Tool}. In: Scarano,
  V., Chiara, R.D., Erra, U. (eds.) Eurographics Italian Chapter Conference.
  The Eurographics Association (2008).
  \doi{10.2312/LocalChapterEvents/ItalChap/ItalianChapConf2008/129-136}

\bibitem{he2016deep}
He, K., Zhang, X., Ren, S., Sun, J.: Deep residual learning for image
  recognition. In: Proceedings of the IEEE conference on computer vision and
  pattern recognition. pp. 770--778 (2016)

\bibitem{horn1987closed}
Horn, B.K.: Closed-form solution of absolute orientation using unit
  quaternions. JOSA A  \textbf{4}(4),  629--642 (1987)

\bibitem{huang2015salicon}
Huang, X., Shen, C., Boix, X., Zhao, Q.: Salicon: Reducing the semantic gap in
  saliency prediction by adapting deep neural networks. In: ICCV (2015)

\bibitem{h36m_pami}
Ionescu, C., Papava, D., Olaru, V., Sminchisescu, C.: Human3.6m: Large scale
  datasets and predictive methods for 3d human sensing in natural environments.
  IEEE Transactions on Pattern Analysis and Machine Intelligence
  \textbf{36}(7),  1325--1339 (jul 2014)

\bibitem{cmrKanazawa18}
Kanazawa, A., Tulsiani, S., Efros, A.A., Malik, J.: Learning category-specific
  mesh reconstruction from image collections. arXiv  (2018)

\bibitem{shapesKarTCM15}
Kar, A., Tulsiani, S., Carreira, J., Malik, J.: Category-specific object
  reconstruction from a single image. In: Computer Vision and Pattern
  Regognition (CVPR) (2015)

\bibitem{kendall2015posenet}
Kendall, A., Grimes, M., Cipolla, R.: Posenet: A convolutional network for
  real-time 6-dof camera relocalization. In: Computer Vision (ICCV), 2015 IEEE
  International Conference on. pp. 2938--2946. IEEE (2015)

\bibitem{lepetit2009epnp}
Lepetit, V., Moreno-Noguer, F., Fua, P.: Epnp: An accurate o (n) solution to
  the pnp problem. International journal of computer vision  \textbf{81}(2),
  ~155 (2009)

\bibitem{li20143d}
Li, S., Chan, A.B.: 3d human pose estimation from monocular images with deep
  convolutional neural network. In: Asian Conference on Computer Vision. pp.
  332--347. Springer (2014)

\bibitem{lin2014microsoft}
Lin, T.Y., Maire, M., Belongie, S., Hays, J., Perona, P., Ramanan, D.,
  Doll{\'a}r, P., Zitnick, C.L.: Microsoft coco: Common objects in context. In:
  European conference on computer vision. pp. 740--755. Springer (2014)

\bibitem{long2014convnets}
Long, J.L., Zhang, N., Darrell, T.: Do convnets learn correspondence? In:
  Advances in Neural Information Processing Systems. pp. 1601--1609 (2014)

\bibitem{lowe2004distinctive}
Lowe, D.G.: Distinctive image features from scale-invariant keypoints.
  International journal of computer vision  \textbf{60}(2),  91--110 (2004)

\bibitem{lu2000fast}
Lu, C.P., Hager, G.D., Mjolsness, E.: Fast and globally convergent pose
  estimation from video images. IEEE Transactions on Pattern Analysis and
  Machine Intelligence  \textbf{22}(6),  610--622 (2000)

\bibitem{mahendran2017joint}
Mahendran, S., Ali, H., Vidal, R.: Joint object category and 3d pose estimation
  from 2d images. arXiv preprint arXiv:1711.07426  (2017)

\bibitem{VNect_SIGGRAPH2017}
Mehta, D., Sridhar, S., Sotnychenko, O., Rhodin, H., Shafiei, M., Seidel, H.P.,
  Xu, W., Casas, D., Theobalt, C.: Vnect: Real-time 3d human pose estimation
  with a single rgb camera. vol.~36 (2017). \doi{10.1145/3072959.3073596},
  \url{http://gvv.mpi-inf.mpg.de/projects/VNect/}

\bibitem{mousavian20173d}
Mousavian, A., Anguelov, D., Flynn, J., Ko{\v{s}}eck{\'a}, J.: 3d bounding box
  estimation using deep learning and geometry. In: Computer Vision and Pattern
  Recognition (CVPR), 2017 IEEE Conference on. pp. 5632--5640. IEEE (2017)

\bibitem{newell2017pixels}
Newell, A., Deng, J.: Pixels to graphs by associative embedding. In: Advances
  in Neural Information Processing Systems. pp. 2168--2177 (2017)

\bibitem{newell2017associative}
Newell, A., Huang, Z., Deng, J.: Associative embedding: End-to-end learning for
  joint detection and grouping. In: Advances in Neural Information Processing
  Systems. pp. 2274--2284 (2017)

\bibitem{newell2016stacked}
Newell, A., Yang, K., Deng, J.: Stacked hourglass networks for human pose
  estimation. In: European Conference on Computer Vision. pp. 483--499.
  Springer (2016)

\bibitem{papadopoulos2017extreme}
Papadopoulos, D.P., Uijlings, J.R., Keller, F., Ferrari, V.: Extreme clicking
  for efficient object annotation. In: 2017 IEEE International Conference on
  Computer Vision (ICCV). pp. 4940--4949. IEEE (2017)

\bibitem{pavlakos20176}
Pavlakos, G., Zhou, X., Chan, A., Derpanis, K.G., Daniilidis, K.: 6-dof object
  pose from semantic keypoints. In: Robotics and Automation (ICRA), 2017 IEEE
  International Conference on. pp. 2011--2018. IEEE (2017)

\bibitem{pavlakos2017coarse}
Pavlakos, G., Zhou, X., Derpanis, K.G., Daniilidis, K.: Coarse-to-fine
  volumetric prediction for single-image 3d human pose. In: Computer Vision and
  Pattern Recognition (CVPR), 2017 IEEE Conference on. pp. 1263--1272. IEEE
  (2017)

\bibitem{Ronchi_2017_ICCV}
Ronchi, M.R., Perona, P.: Benchmarking and error diagnosis in multi-instance
  pose estimation. In: The IEEE International Conference on Computer Vision
  (ICCV) (Oct 2017)

\bibitem{simonyan2014very}
Simonyan, K., Zisserman, A.: Very deep convolutional networks for large-scale
  image recognition. arXiv preprint arXiv:1409.1556  (2014)

\bibitem{su2015render}
Su, H., Qi, C.R., Li, Y., Guibas, L.J.: Render for cnn: Viewpoint estimation in
  images using cnns trained with rendered 3d model views. In: Proceedings of
  the IEEE International Conference on Computer Vision. pp. 2686--2694 (2015)

\bibitem{szeto2017click}
Szeto, R., Corso, J.J.: Click here: Human-localized keypoints as guidance for
  viewpoint estimation. arXiv preprint arXiv:1703.09859  (2017)

\bibitem{taylor2012vitruvian}
Taylor, J., Shotton, J., Sharp, T., Fitzgibbon, A.: The vitruvian manifold:
  Inferring dense correspondences for one-shot human pose estimation. In:
  Computer Vision and Pattern Recognition (CVPR), 2012 IEEE Conference on. pp.
  103--110. IEEE (2012)

\bibitem{tompson2015efficient}
Tompson, J., Goroshin, R., Jain, A., LeCun, Y., Bregler, C.: Efficient object
  localization using convolutional networks. In: Proceedings of the IEEE
  Conference on Computer Vision and Pattern Recognition. pp. 648--656 (2015)

\bibitem{tompson2014joint}
Tompson, J.J., Jain, A., LeCun, Y., Bregler, C.: Joint training of a
  convolutional network and a graphical model for human pose estimation. In:
  Advances in neural information processing systems. pp. 1799--1807 (2014)

\bibitem{toshev2014deeppose}
Toshev, A., Szegedy, C.: Deeppose: Human pose estimation via deep neural
  networks. In: Proceedings of the IEEE conference on computer vision and
  pattern recognition. pp. 1653--1660 (2014)

\bibitem{tulsiani2015pose}
Tulsiani, S., Carreira, J., Malik, J.: Pose induction for novel object
  categories. In: Proceedings of the IEEE International Conference on Computer
  Vision. pp. 64--72 (2015)

\bibitem{tulsiani2015viewpoints}
Tulsiani, S., Malik, J.: Viewpoints and keypoints. In: Proceedings of the IEEE
  Conference on Computer Vision and Pattern Recognition. pp. 1510--1519 (2015)

\bibitem{drcTulsiani17}
Tulsiani, S., Zhou, T., Efros, A.A., Malik, J.: Multi-view supervision for
  single-view reconstruction via differentiable ray consistency. In: Computer
  Vision and Pattern Regognition (CVPR) (2017)

\bibitem{DBLP:conf/cvpr/WeiHCVL16}
Wei, L., Huang, Q., Ceylan, D., Vouga, E., Li, H.: Dense human body
  correspondences using convolutional networks. In: 2016 {IEEE} Conference on
  Computer Vision and Pattern Recognition, {CVPR} 2016, Las Vegas, NV, USA,
  June 27-30, 2016. pp. 1544--1553 (2016)

\bibitem{wei2016convolutional}
Wei, S.E., Ramakrishna, V., Kanade, T., Sheikh, Y.: Convolutional pose
  machines. In: Proceedings of the IEEE Conference on Computer Vision and
  Pattern Recognition. pp. 4724--4732 (2016)

\bibitem{marrnet}
Wu, J., Wang, Y., Xue, T., Sun, X., Freeman, W.T., Tenenbaum, J.B.: {MarrNet:
  3D Shape Reconstruction via 2.5D Sketches}. In: Advances In Neural
  Information Processing Systems (2017)

\bibitem{wu2016single}
Wu, J., Xue, T., Lim, J.J., Tian, Y., Tenenbaum, J.B., Torralba, A., Freeman,
  W.T.: Single image 3d interpreter network. In: European Conference on
  Computer Vision. pp. 365--382. Springer (2016)

\bibitem{xiang2016objectnet3d}
Xiang, Y., Kim, W., Chen, W., Ji, J., Choy, C., Su, H., Mottaghi, R., Guibas,
  L., Savarese, S.: Objectnet3d: A large scale database for 3d object
  recognition. In: European Conference on Computer Vision. pp. 160--176.
  Springer (2016)

\bibitem{xiang2014beyond}
Xiang, Y., Mottaghi, R., Savarese, S.: Beyond pascal: A benchmark for 3d object
  detection in the wild. In: Applications of Computer Vision (WACV), 2014 IEEE
  Winter Conference on. pp. 75--82. IEEE (2014)

\bibitem{xu2015show}
Xu, K., Ba, J., Kiros, R., Cho, K., Courville, A., Salakhudinov, R., Zemel, R.,
  Bengio, Y.: Show, attend and tell: Neural image caption generation with
  visual attention. In: International Conference on Machine Learning. pp.
  2048--2057 (2015)

\bibitem{yang2017learning}
Yang, W., Li, S., Ouyang, W., Li, H., Wang, X.: Learning feature pyramids for
  human pose estimation. In: The IEEE International Conference on Computer
  Vision (ICCV). vol.~2 (2017)

\bibitem{yi2016scalable}
Yi, L., Kim, V.G., Ceylan, D., Shen, I., Yan, M., Su, H., Lu, C., Huang, Q.,
  Sheffer, A., Guibas, L., et~al.: A scalable active framework for region
  annotation in 3d shape collections. ACM Transactions on Graphics (TOG)
  \textbf{35}(6), ~210 (2016)

\bibitem{yuan20173d}
Yuan, S., Garcia-Hernando, G., Stenger, B., Moon, G., Chang, J.Y., Lee, K.M.,
  Molchanov, P., Kautz, J., Honari, S., Ge, L., et~al.: 3d hand pose
  estimation: From current achievements to future goals. arXiv preprint
  arXiv:1712.03917  (2017)

\bibitem{zhang2014part}
Zhang, N., Donahue, J., Girshick, R., Darrell, T.: Part-based r-cnns for
  fine-grained category detection. In: European conference on computer vision.
  pp. 834--849. Springer (2014)

\bibitem{zhou2016learning}
Zhou, T., Krahenbuhl, P., Aubry, M., Huang, Q., Efros, A.A.: Learning dense
  correspondence via 3d-guided cycle consistency. In: Proceedings of the IEEE
  Conference on Computer Vision and Pattern Recognition. pp. 117--126 (2016)

\bibitem{Zhou_2017_ICCV}
Zhou, X., Huang, Q., Sun, X., Xue, X., Wei, Y.: Towards 3d human pose
  estimation in the wild: A weakly-supervised approach. In: The IEEE
  International Conference on Computer Vision (ICCV) (Oct 2017)

\bibitem{zhou2017unsupervised}
Zhou, X., Karpur, A., Gan, C., Luo, L., Huang, Q.: Unsupervised domain
  adaptation for 3d keypoint prediction from a single depth scan. arXiv
  preprint arXiv:1712.05765  (2017)

\bibitem{zhou2016deep}
Zhou, X., Sun, X., Zhang, W., Liang, S., Wei, Y.: Deep kinematic pose
  regression. arXiv preprint arXiv:1609.05317  (2016)

\bibitem{zhou2016model}
Zhou, X., Wan, Q., Zhang, W., Xue, X., Wei, Y.: Model-based deep hand pose
  estimation. arXiv preprint arXiv:1606.06854  (2016)

\end{thebibliography}

\clearpage
\section{Supplementary Material}

\subsection{Pose Induction on Pascal3D+}

\begin{table}[t]
\scriptsize
\center{
\begin{tabular}{l|cc}
  \hline
  & Motorcycle & Bus \\ \hline		
$\mathit{Acc}_\frac{\pi}{6}$ (Similar Classifier Transfer~\cite{tulsiani2015viewpoints}) & 0.58 & 0.50 \\ 
$\mathit{Acc}_\frac{\pi}{6}$ (General Classifier~\cite{tulsiani2015viewpoints}) & 0.55 & 0.80  \\
$\mathit{Acc}_\frac{\pi}{6}$ (General Classifier Res18) & 0.58 & 0.79  \\
$\mathit{Acc}_\frac{\pi}{6}$ (Ours) & 0.55 & 0.63 
\\ \hline 
\end{tabular}
}
{\caption{Viewpoint estimation of novel categories on Pascal3D+~\cite{xiang2014beyond}. We compare with the baselines from Tulsiani et al.~\cite{tulsiani2015pose} and our re-trained ResNet18~\cite{he2016deep} model. The results are shown in $\mathit{Acc}_\frac{\pi}{6}$.}
\label{table:pascal_novel}}
\vspace{-0.6cm}
\end{table}

The pose induction for novel categories problem has been studied by Tulsiani et al.~\cite{tulsiani2015pose}.
They proposed two baselines for viewpoint induction: i) Similar Classifier Transfer (SCT), which uses the viewpoint classifier of a manually defined similar category for the novel category (e.g., use bicycle classifier for motorcycle); 
2) General Classifier (GC),  which trains a category-agnostic viewpoint classifier (similar to our Res18-General baseline in Table. 2 of our main paper). 
For evaluation, they~\cite{tulsiani2015pose} exclude two categories (Motorcycle and Bus) from the Pascal3D+ training set~\cite{xiang2014beyond} and evaluate viewpoint estimation on these two categories with the same protocol of~\cite{tulsiani2015viewpoints}. 
We compare our proposed method on viewpoint estimation with their baselines in Table~\ref{table:pascal_novel}. 

Our keypoint alignment-based viewpoint estimator achieved lower performance than direct general viewpoint classification. This can be understood from the following factors. First, the viewpoint estimation task has shown itself not to be category-specific. As shown in Table. 2 of the main paper, Res18-General has a very close performance with Res18-Specific ($\mathit{Acc}_\frac{\pi}{6}$ 0.79 vs. 0.81), indicating that viewpoint estimation does not benefit a lot from category-specific design.
However, keypoint estimation is inherently category-specific, and keypoint definitions vary widely per category.
Our system places emphasis on learning the geometry of each training category, and such information is only weakly connected to the viewpoint estimation task.
Despite these limitations, our keypoint-based method is able to achieve encouraging results on pose induction ($55\%$ accuracy on Motorcycle, $62\%$ accuracy on Bus). Moreover, as indicated in the main paper, the view-point estimation performance of our method is highly correlated with the consistency of keypoint predictions and CanViewFeature. On novel categories, they become less consistent, leading to a drop in viewpoint estimation accuracy. However, one can certainly employ domain adaptation techniques to improve their consistency. We leave this as a direction for future research.

Our proposed method is currently the only learning-based method to induct keypoint estimation to novel categories. However, we remark that we avoid directly evaluating keypoint localization performance, as keypoint detection task on novel category is ill-posed. Keypoint definitions are subjective on novel objects, e.g., our method consistently predicts frontal lights as keypoint for bus, while the annotations of Pascal3D+~\cite{xiang2014beyond} do not, presumably due to light being defined as a keypoint on a car but not on a bus.

\subsection{ObjectNet3D dataset split}

\begin{table*} [t]
 \begin{center} 
 \tiny
 \begin{tabular}{c|c|c|c|c|c|c} 
 \hline
aeroplane & camera & eraser & jar & pencil & shovel & toothbrush   \\
ashtray & can & eyeglasses & kettle & piano & sign & train   \\
backpack & cap & fan & key & pillow & skate & trash bin   \\
basket & car & faucet & keyboard & plate & skateboard & trophy   \\
\underline{bed} & \underline{cellphone} & \underline{filing cabinet} & \underline{knife} & \underline{pot} & \underline{slipper} & \underline{tub}   \\
bench & chair & fire extinguisher & laptop & printer & sofa & tvmonitor   \\
bicycle & clock & fish tank & lighter & racket & speaker & vending machine   \\
blackboard & coffee maker & flashlight & mailbox & refrigerator & spoon & washing machine   \\
boat & comb & fork & microphone & remote control & stapler & watch   \\
\underline{bookshelf} & \underline{computer} & \underline{guitar} & \underline{microwave} & \underline{rifle} & \underline{stove} & \underline{wheelchair}   \\
bottle & cup & hair dryer & motorbike & road pole & suitcase &    \\
bucket & desk lamp & hammer & mouse & satellite dish & teapot &    \\
bus & diningtable & headphone & paintbrush & scissors & telephone &    \\
cabinet & dishwasher & helmet & pan & screwdriver & toaster &    \\
\underline{calculator} & \underline{door} & \underline{iron} & \underline{pen} & \underline{shoe} & \underline{toilet} &    \\
\hline
\end{tabular} 
{\caption{List of categories on ObjectNet3D~\cite{xiang2016objectnet3d}. The novel categories (only used for testing) is shown in underline.}
\label{table:cat}}
\end{center} 

\end{table*}

The detailed training and testing categories split is shown in Table.~\ref{table:cat}.
ObjectNet3D~\cite{xiang2016objectnet3d} contains about 50k training samples in total, but only 20k of them have keypoint annotations. 
We use the 20k subset of the training set for training and the validation set for testing. In total, we collected 19k images for training, and 4k images for novel categories.

\begin{table}[t]
\scriptsize
\center{
\begin{tabular}{l|*{7}{c}|r}
  \hline
  & Head & Shoulder & Elbow & Wrist & Hip & Knee  & Ankle & Total \\ 	
  \hline
  HourglassNetwork w.~\cite{newell2016stacked} oracle ID & 97.44 & 98.27 & 94.02 & 92.22 & 93.30 & 90.49 & 86.02 & 93.22 \\ 
  StarMap with oracle Id & 92.12 & 93.65 & 90.49 & 86.09 & 82.40 & 87.23 & 82.22 & 88.17 \\
  \hline 
  HourglassNetwork~\cite{newell2016stacked}& 96.49 &95.38 &89.16 &84.89 & 87.73 &84.08 & 80.30 & 88.39 \\ 
  StarMap with learned Id & 91.00 & 88.69 & 83.02 & 73.58 & 74.16 & 76.67 & 69.01 & 79.85 \\ 
   \hline 
\end{tabular}
}{\caption{Results on MPII. The results are shown in PCKh@0.5, which is the percentage of correct keypoint whose diviation are within 0.5 of head bounding box.}
\vspace{-0.0cm}
\label{table:mpii-pckh5}}
\end{table}
\begin{table}[t]
\tiny
\centering
\renewcommand{\tabcolsep}{1pt}
\begin{tabular}{l|c|c|c|c|c|c|c|c|c|c|c|c|c|c|c|c}
\hline
                &        &         &        &       &       &        &        &         & Sit   &       & Take  &       &       & Walk  & Walk  & \multicolumn{1}{l}{} \\
\multicolumn{1}{c|}{Method}                      & Direct & Discuss & Eat & Greet & Phone & Pose & Purch. & Sit & Down  & Smoke & Photo & Wait  & Walk  & Dog   & Pair  & All                  \\ \hline 
Mehta~\cite{VNect_SIGGRAPH2017}  &62.6 & 78.1 & 63.4 & 72.5 & 88.3 & 63.1 & 74.8 & 106.6 & 138.7 & 78.8 & 93.8 & 73.9 & 55.8 & 82.0& 59.6 & 80.5\\
Zhou~\cite{Zhou_2017_ICCV}     & 54.8   & 60.7   & 58.2   & 71.4 & 62.0 & 53.8  & 55.6  & 75.2 & 111.6 & 64.2 & 65.5  &  66.1 & 63.2  & 51.4 & 55.3  & 64.9                \\
Ours  & 56.6    &  62.6    &     54.7 & 64.4   & 69.7 & 53.0  & 54.3    & 80.7 &122.4  & 65.3 &69.6  & 57.9  & 47.0 & 65.0  & 52.5  & 65.77 \\
\hline
\end{tabular}
{\caption{Results on H36M~\cite{h36m_pami} Dataset. The results are shown in Mean Per Joint Position Error (in mm).}\label{Table:H36M}}

\vspace{-0.3cm}
\end{table}

\subsection{Human pose estimation.} 
In the main paper we have considered evaluating our approach on rigid objects. We show that the results are consistent on a different task, namely, human pose estimation. 

\subsubsection{2D human pose estimation}
We first evaluate StarMap on the task of 2D human pose estimation on the MPII Dataset~\cite{andriluka14cvpr}, by replacing the 16-channel output of state-of-the-art HourglassNet~\cite{newell2016stacked} with a one-channel StarMap and a two-channel 2D canonical feature. 
As shown in Table~\ref{table:mpii-pckh5}, our method leads to encouraging results when compared to the default HourglassNet~\cite{newell2016stacked}, especially when assigned oracle identification, which means we can see very similar visual results by using 1 output channel instead of 16.

\begin{figure}[t]
\centering
\includegraphics[width=0.9\linewidth]{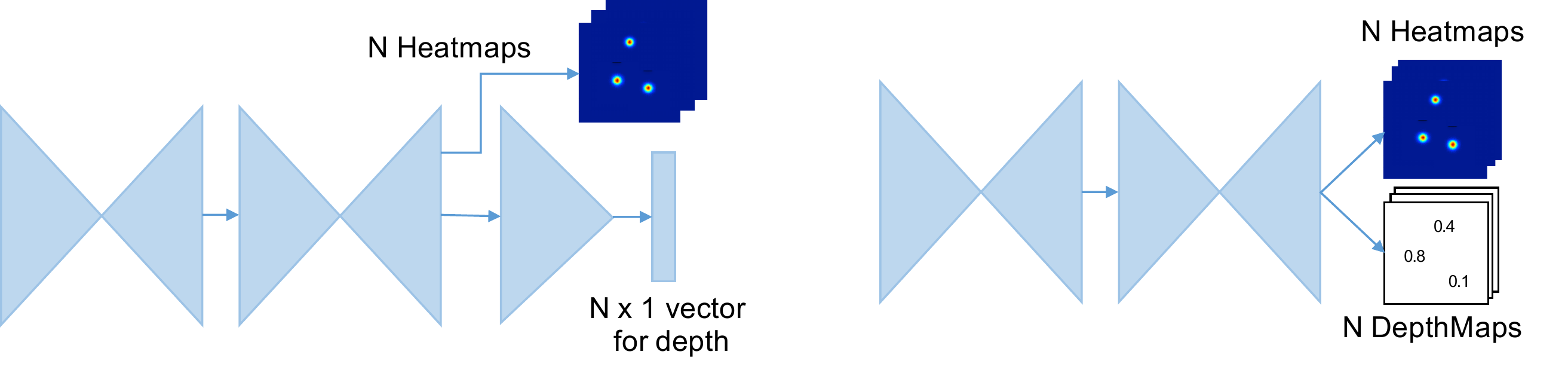}
\caption{Difference between our depth regression module and Zhou et al.~\cite{Zhou_2017_ICCV}. Left:~\cite{Zhou_2017_ICCV} architecture, which uses a \emph{sub-network} for depth regression. Right: Ours architecture, which uses N additional \emph{channels} for depth regression.}
\label{fig:zhou}
\vspace{-0.3cm}
\end{figure}

\subsubsection{3D human pose estimation}
The \emph{DepthMap} representation, which associate each 2D joint with a depth value in a \emph{map representation}, 
can be a simplified 3D keypoint representation. 
It is contrast to Zhou et al.~\cite{Zhou_2017_ICCV} who represent 3D keypoint as 2D heatmap and depth vector learned with an additional subnetwork.
More specifically, Zhou et al.~\cite{Zhou_2017_ICCV} proposes to decouple the $(x, y, z)$ 3D coordinate into $(u, v)$ image coordinate and depth $d$ (see our Section. 3.2) in a weak-perspective camera model,
which enables using rich 2D in-the-wild data~\cite{andriluka14cvpr} in training. 
For estimating the depth $d$ of each joint, they use an additional depth regression \emph{sub-network} on the top of the 2D network, which is cumbersome (i.e., introducing more hyper-parameters for designing the sub-network and increasing the feed forward time). 
When using our DepthMap encoding, which augments the $(u, v)$ heatmaps with an additional depth \emph{channel} and associates the depth value on the heatmap peak location, 
we can replace the sub-network~\cite{Zhou_2017_ICCV} with channels.
We illustrate the difference in Fig.~\ref{fig:zhou}.

We evaluate it by replacing the regression subnetwork of~\cite{Zhou_2017_ICCV} with an N-channel \emph{DepthMap} for 3D human pose estimation on Human 3.6M dataset~\cite{h36m_pami}. 

Human3.6M dataset~\cite{h36m_pami}, which contains about 3.6 millions frames of images, each with accurate 3D human joint location annotations. Following~\cite{Zhou_2017_ICCV,pavlakos2017coarse}, both training and testing are done on a $5 \times$ down-sampled subset. 
We follow the standard protocol to use 5 subjects for training and 2 subjects for testing. 
The error is measured in mean per joint position error (MPJPE) in millimeters after aligning the root joint location with ground truth and assuming a fixed average scale~\cite{Zhou_2017_ICCV,pavlakos2017coarse}.
All the experiment settings are the same with~\cite{Zhou_2017_ICCV}.

The results in Table.~\ref{Table:H36M} show that our DepthMap representation achieves very close performance with the original design of Zhou et al.~\cite{Zhou_2017_ICCV}, 
while saving about $1/5$ network parameter (from the depth-regression sub-network).
We also compare with Mehta et al.~\cite{VNect_SIGGRAPH2017}, who also use a map representation for $3D$ coordinates. 
Instead of directly using the $(u, v)$ coordinate from 2D heatmap (with a weak-perspective camera model), they regress the full $(x, y, z)$ coordinates at the peak heatmap location with a full-perspective camera model. 
Also, they use a modified ResNet50~\cite{he2016deep} architecture instead of HourglassNetwork~\cite{newell2016stacked}. 
Our results are considerably better than theirs, showing the effectiveness of the decoupled weak-perspective 3D keypoint representation.

\subsection{More Qualitative Results}
This section is removed due to arXiv size limit. Please visit the project page (\url{https://github.com/xingyizhou/StarMap}) for more qualitative results.

\end{document}